\documentclass{article}
\usepackage{amsmath,amssymb}
\usepackage{subcaption}
\usepackage{bm}
\usepackage{graphicx}
\usepackage{natbib}

     \usepackage[preprint]{neurips_2021}

\usepackage[utf8]{inputenc} % allow utf-8 input
\usepackage[T1]{fontenc}    % use 8-bit T1 fonts
\usepackage{hyperref}       % hyperlinks
\usepackage{url}            % simple URL typesetting
\usepackage{booktabs}       % professional-quality tables
\usepackage{amsfonts}       % blackboard math symbols
\usepackage{nicefrac}       % compact symbols for 1/2, etc.
\usepackage{microtype}      % microtypography
\usepackage{xcolor}         % colors
\usepackage{amsthm, amsmath}

\title{GNisi: A graph network for reconstructing Ising models from multivariate binarized data}

\author{%
  Emma Slade\\%\thanks{Use footnote for providing further information about author (webpage, alternative address)---\emph{not} for acknowledging funding agencies.} \\
  GSK.ai,  London, \\N1C 4AG, UK\\
  \texttt{emma.x.slade@gsk.com} \\
  \And
  Sonya Kiselgof \\ GSK.ai, 25 Basel St, \\Petah Tikvah, Israel \\ \texttt{sofya.x.kiselgof@gsk.com}\\
  \And
  Lena Granovsky \\ GSK.ai, 25 Basel St, \\Petah Tikvah, Israel \\ \texttt{lena.x.granovsky@gsk.com}\\
  \And
  Jeremy L. England\\
School of Physics, Georgia Institute of Technology, Atlanta, GA, USA \\ GSK.ai, 25 Basel St, Petah Tikvah, Israel \\
  \texttt{jeremy.l.england@gsk.com} \\
}

\newcommand{\EE}{\mathcal{E}}
\newcommand{\GG}{\mathcal{G}} 
\newcommand{\VV}{\mathcal{V}} 
\newcommand{\NN}{\mathcal{N}} 
\newcommand{\R}{\mathbb{R}}

\newcommand{\vv}{\boldsymbol{v}}
\newcommand{\ee}{\boldsymbol{e}}
\newcommand{\xx}{\boldsymbol{x}}

\newcommand{\yy}{\boldsymbol{y}}

\newcommand{\ie}{\emph{i.e.}}
\newcommand{\eg}{\emph{e.g.}}

\hypersetup{draft}

\begin{document}

\maketitle
\maketitle

\begin{abstract}
Ising models are a simple generative approach to describing interacting binary variables. They have proven useful in a number of biological settings because they enable one to represent observed many-body correlations as the separable consequence of many direct, pairwise statistical interactions. The inference of Ising models from data can be computationally very challenging and often one must be satisfied with numerical approximations or limited precision. In this paper we present a novel method for the determination of Ising parameters from data, called GNisi, which uses a Graph Neural network trained on known Ising models in order to construct the parameters for unseen data. We show that GNisi is more accurate than the existing state of the art software, and we illustrate our method by applying GNisi to gene expression data.

\end{abstract}

\section{Introduction}

The ability to infer graphical models which effectively model interactions in a dataset is a powerful numerical tool across many scientific disciplines. The Ising model, as it is known in the language of statistical physics, is a particularly well-known approach which can be used to understand the underlying correlations in any form of binary data (“spins”) via pairwise and single-site interactions. Inferring the parameters of an Ising model from data, however, is computationally demanding and often intractable, as it is an inverse problem that typically relies on sampling from the Boltzmann distribution the Ising parameters imply.

Existing techniques to solve the inverse Ising model approximately include mean-field inference~\citep{10.1162/089976698300017386,coniii}, pseudo-likelihood~\citep{PhysRevLett.108.090201,10.1214/09-AOS691}, and perturbative expansion methods~\citep{Nguyen_2012} which can capture the general form of the graph.  More accurate methods include Monte Carlo simulation~\citep{mc1, 10.1371journal.pcbi.1003776, ACKLEY1985147} and the adaptive cluster expansion method~\citep{ace}, which, while more accurate may be slow to converge or numerically unstable.

Given the utility of inverse Ising models for deriving direct interactions from observed covariations in large numbers of variables, it is natural that there are have been many successful applications in the biological domain. Such approaches have been used for determining protein structure~\citep{Obermayer2014, 10.1371/journal.pone.0028766, Qin690, Shakhnovich7195}, viral fitness in HIV~\citep{10.1371journal.pcbi.1003776, Barton1965}, correlated states in neural populations~\citep{Schneidman2006}, correlations in nucleotide sequences~\citep{10.1371/journal.pcbi.0030231, 10.1093/bioinformatics/btn313} as well as modelling a statistical description of gene expression levels~\citep{Lezon19033}.

Graph networks are a deep learning architecture designed to learn from graph-structured data. They were originally proposed in~\citep{gori2005new,scarselli2008graph}  (see \eg,~\citep{battaglia2018relational,hamilton2020graph,wu2020comprehensive} for an overview). The application of graph networks to predict the behavior of complex physical systems has succeeded in a wide range of settings~\citep{pmlr-v119-sanchez-gonzalez20a, pfaff2021learning}, including the modelling of spin-glass~\citep{spinglass}. 
In this paper we present a novel method for solving inverse Ising problems using graph networks, which we denote GNisi, which alleviates the slow convergence or inaccuracy of previous methods. In our model, the graph network is trained on known Ising models, generated using Monte Carlo methods, and then the trained network is used to determine the Ising parameters for unseen binary data.

We will firstly compare our method to existing methods for solving inverse Ising problems on a generated dataset against which we can compare to ground truth. We then apply our method to genomic data from the Cancer Cell Line Encyclopaedia project (CCLE)~\citep{Ghandi2019,Li2019,Barretina2012,Stransky2015} under CC BY 4.0 licence~\footnote{\url{https://creativecommons.org/licenses/by/4.0/legalcode}}, in order to model the covariation of gene expression from the anonymized~\citep{depmap_2020} dataset, for two subsets of genes that are known to be highly correlated.

\section{Theoretical outline}
\subsection{The Ising model} \label{sec:ising-model}
For a dataset of bits, $\xx = \left[ x_1 , \dots, x_n \right] $, the Ising model for that data is given by the probability distribution
\begin{equation}
p(\xx) = \frac{1}{Z} \exp\left(- \sum_i  h_i x_i - \sum_{i < j} u_{ij} x_i x_j \right) \label{eq:ising_distribution} \,,
\end{equation}
where the partition function is the normalisation such that probabilities sum to  identity
\begin{equation}
Z = \sum_i \exp^{ \beta H_i} \label{eq:partition_function} \,,
\end{equation}
where $\beta$ is the inverse temperature, and the Hamiltonian is defined as
\begin{equation}
H =- \sum_i  h_i x_i - \sum_{i < j} u_{ij} x_i x_j  \label{eq:hamiltonian} \,.
\end{equation}
One can show the probability distribution, Eq.~\eqref{eq:ising_distribution} maximizes the Shannon entropy
\begin{equation}
S = - \sum_{\xx} p(\xx) \log p(\xx) \label{eq:shannon_entropy} \,,
\end{equation}
under the constraint that the first and second moments of $\xx$ are appropriately defined. In particular, we require the sample means and covariances to match the connected first and second moments
\begin{equation} \label{eq:first_moment}
\langle x_i \rangle = \sum_{\xx} p(\xx) x_i 
\end{equation}
\begin{equation} \label{eq:second_moment}
\langle x_i x_j \rangle - \langle x_i \rangle \langle x_j \rangle= \sum_{\xx} p(\xx) x _i x_j  \,.
\end{equation}
The problem is then to find the parameters $u_{ij}, h_i$, such that the first and second moments match the mean values of $x_i$ and $x_i x_j$ observed in the data. A much simpler task is to compute the pairwise term, $u_{ij}$ by inverting the covariance matrix of the dataset, $\text{cov}^{-1}(x_i, x_j)$, if it is invertible for the system. One may show analytically that, in the limit of high temperature, the two matrices are equivalent, however, there are many situations in which the data distribution will not be well-described by such a Gaussian form.

\subsection{Graph neural networks}
\begin{figure}[h!] 
  \centering
  \begin{subfigure}[h]{0.2\textwidth}
    \centering
    \includegraphics[width=0.9\textwidth]{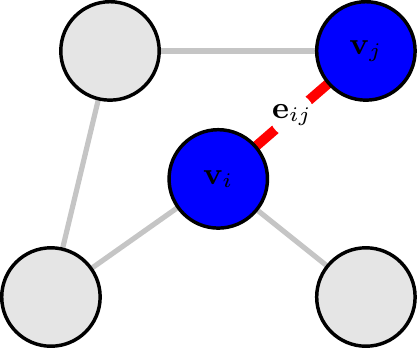}
    \caption{Edge update}
  \end{subfigure}
    \begin{subfigure}[h]{0.2\textwidth}
    \centering
    \includegraphics[width=0.9\textwidth]{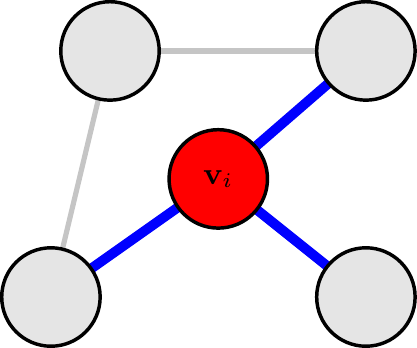}
    \caption{Node update}
  \end{subfigure}
      \caption{Illustration of graph network updates.  In red the object being updated, and in blue the quantities used to perform the update.} \label{fig:gnns}
\end{figure}

Let a graph be defined as $\GG=(\VV,\EE)$ where $\VV=\{1,\dots, N\}$ is the set of nodes and $\EE\subseteq \VV\times\VV$ is the set of edges connecting nodes in $\VV$. We denote $\NN_i$ as the set of neighbours of node $i$. Using as similar notation as in~\citep{battaglia2018relational}, we let $\vv_i\in\R^{n_v}$ represent the attributes of node $i$ for all $i\in\VV$ and $\ee_{ij}\in\R^{n_e}$ represent the attributes of edge $(i,j)$ for all $(i,j)\in\EE$. We illustrate the architecture in Fig.~\ref{fig:gnns}.
Then, a graph network can be characterized in terms of edge and node updates as
\begin{subequations}\label{eq:gnn}
\begin{align}
  \ee_{ij}^+ &= \phi^e\big(\ee_{ij}, \vv_i, \vv_j\big),&\forall (i,j)\in\EE\\
  \vv_i^+ &= \phi^v\big(\rho^{e\to v}\big(\{\ee_{ij}^+\}_{j\in\NN_i}\big), \vv_i \big),&\forall i\in\VV
\end{align}
\end{subequations}
where $\phi^e:\R^{n_e+ 2n_v }\to \R^{n_e}$, $\phi^v:\R^{n_e+ n_v}\to \R^{n_v}$, are the update functions to be learned and $\rho^{e\to v}$ is an aggregation function reducing a set of elements to a single one via some input's permutation equivariant transformation.
In the particular case of Ising models, we are working with undirected graphs. If, as we do here, one wishes to infer the existence of the edges (\ie\, determine whether two bits are correlated), then we must start with a fully-connected graph, where $\NN_i=\VV\setminus \{i\} \forall i$.
\section{Methods}
\subsection{Metropolis Monte Carlo simulations}
In this paper we aim to learn the Ising model parameters, $h_i, u_{ij}$ of an unseen system by training a graph network on a wide range of known Ising models. 
To generate the Ising models, we use the Metropolis Monte Carlo algorithm~\citep{metropolismc}, which calculates valid bit-strings for an Ising model using single-spin flip dynamics. In each Monte Carlo iteration of the algorithm, a single spin is selected at random and the change in the  energy of the system, $\Delta E$ is calculated as the difference between the Hamiltonian~\eqref{eq:hamiltonian} for the two systems. If $\Delta E \le 0$, the change is favourable and the spin is flipped. Otherwise, the spin is flipped only if $\text{e}^{- \beta \Delta E} > x$, where $x$ is a random number in the range $[0,1]$. This process is repeated for every site in the lattice, until the model has converged and the energy can no longer decrease.

\begin{figure}[h!] 
  \centering
  \begin{subfigure}[h]{0.33\textwidth}
    \centering
    \includegraphics[width=0.9\textwidth]{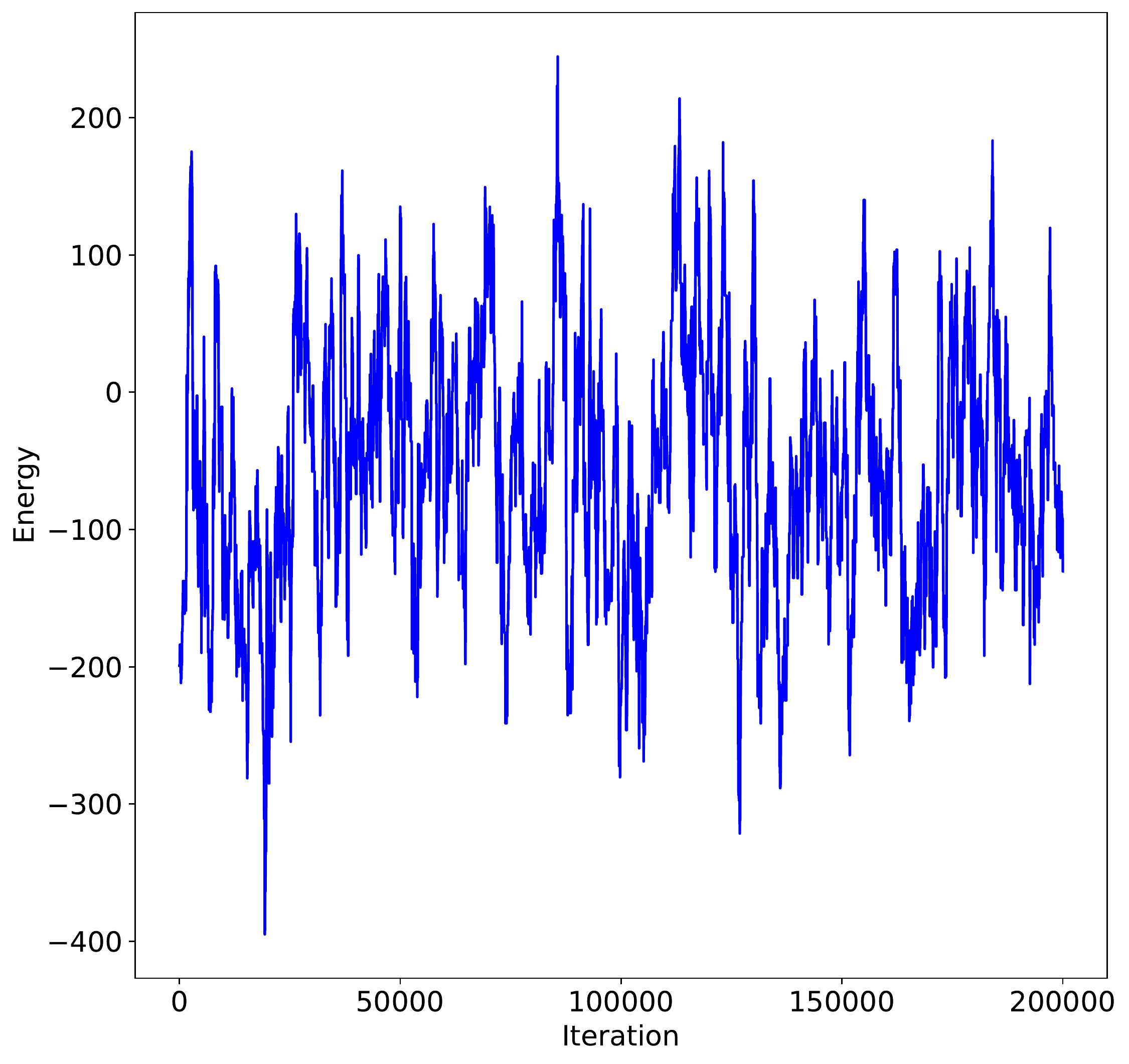}
%    \caption{Convergence of the energy for an Ising model at high temperature}
  \end{subfigure}
    \begin{subfigure}[h]{0.33\textwidth}
    \centering
    \includegraphics[width=0.9\textwidth]{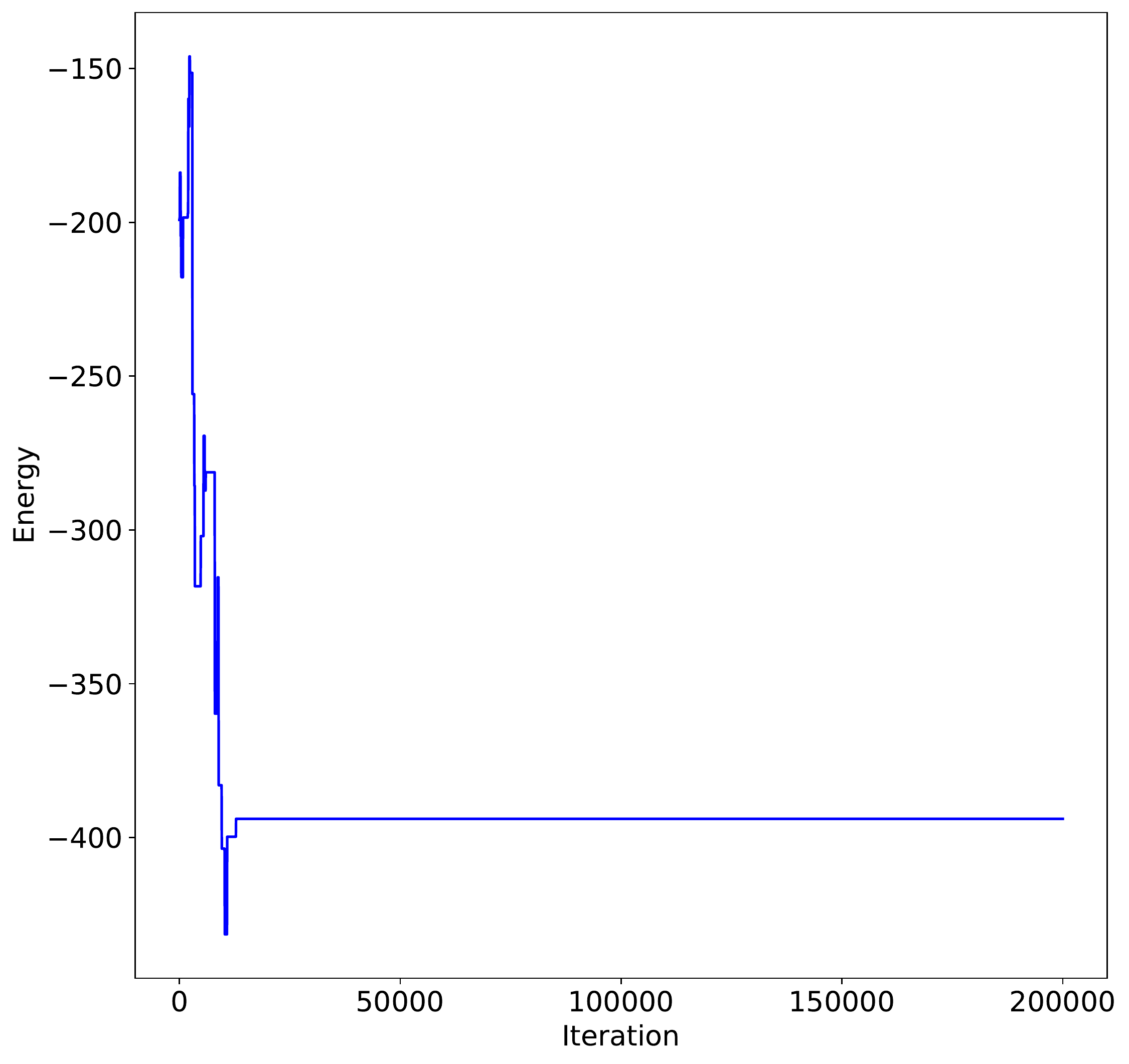}
%    \caption{Convergence of the energy for an Ising model at low temperature}
  \end{subfigure}
      \caption{Convergence of the energy for Ising models at high (left) and low (right) temperatures.} \label{fig:mc_metropolis}
\end{figure}

In Fig.~\ref{fig:mc_metropolis} we show representative training runs for 2 Ising models at low and high temperatures. We observe that the energy of the high temperature model (left) does not converge, as one would expect, whereas the low temperature system (right) converges rapidly.

In order to allow the network to learn a wide range of Ising models, we generate the training data for a wide range of temperatures, lattice sizes and sparsities.
The Monte Carlo data is converted to fully-connected graphs for the purposes of training; the initial node features of each graph are the states of the spins $(0,1)$, whilst the edge attributes are the value of the coupling determined from the Ising model parameters.

\subsection{Training procedure}
For each Ising model we present the network with 1000 samples of converged spins generated using the Metropolis Monte Carlo algorithm, split between 800 samples in training and 200 validation.
We minimize the $\mathcal{L}_2$ loss between the predicted and true Ising parameters and we use early-stopping, selecting the model with the highest accuracy on the validation set.  The test set consists of entirely unseen models, each with 1000 samples.

The full graph network is composed of 6 graph layers, with the node and edge encoding each composed of 1 hidden layer with dimensionality [123,119,28,126,126,126]. The ReLU activation function is used and we use the Adam optimizer~\citep{kingma2014method} with a fixed learning rate of $1e^{-3}$. The number of nodes in the hidden layers as well as number of graph layers, the learning rate and optimizer are determined by optimising the validation loss of the models with $\beta = 1$ and 50\% sparsity using the Optuna~\citep{optuna_2019} framework.
Within each graph layer, each node is updated based on its previous embedding and the sum of the incoming edges, whilst the edges are updated based on their previous embeddings and the embeddings of the connected nodes. 

\section{Experimental results}

In this section we will show the results of the GNisi framework on two separate datasets. Firstly, we will discuss how GNisi performs on a test set composed of unseen Ising models with differing physical properties where ground truth is known. We will additionally compare with existing codes which solve the inverse Ising task using adaptive cluster expansion~\citep{ace, coniii}, which are the current state of the art in solving this class of problems. Secondly we apply GNisi to data from the Cancer Cell Line Encyclopaedia project (CCLE)~\citep{Ghandi2019,Li2019,Barretina2012,Stransky2015}, which consists of 1376 cell lines, in order to model the covariation of expression of a set of highly correlated genes.

\subsection{Results on unseen Ising models} \label{sec:unseen}

Quantifying how well we are able to generate an accurate Ising model for a set of bit-strings can be done in several ways. Matching the ground truth matrix matrix-element by matrix-element can of course only be assessed when one has ground truth to compare with, which is not the case in general. As discussed in Sec.~\ref{sec:ising-model}, in the maximum entropy approach, the aim is to match the first and second moments of $\xx$ such that the Shannon entropy of the system is maximized. Here we will aim to show that with GNisi we are able to additionally match the sampled third order moments of $\xx$
\begin{equation} \label{eq:third_moment}
\langle x_i x_j x_k \rangle = \sum_{\xx} p(\xx) x _i x_j x_k \,,
\end{equation}
which provides us with much higher level of precision of how closely the generated Ising model matches the ground truth data. Furthermore, we will show that we are able to closely match the underlying Boltzmann distribution of the data, Eq.~\eqref{eq:partition_function}. We note that, in order to prevent any additional overfitting on the test set, the bit-string samples used to construct the Ising model, which are generated from the Metropolis Monte Carlo algorithm, are not used in quantifying the goodness of fit. In other words, we generate a distinct set of bit-string samples when constructing the Boltzmann distributions.

We compare our results for a representative model from the test set, namely an Ising model of size 50, at a low temperature and 25\% sparsity in the matrix elements. We can compare the Ising model obtained by GNisi with the ones obtained with the two current open source packages for solving inverse Ising models;  ACE v.1~\citep{ace} and coniii v.2.3.0~\citep{coniii}, both released under MIT licence. In order to provide a fair comparison, we run the packages using default settings when possible, in order to prevent fine-tuning the results on a model-by-model basis as there is no equivalent to fine-tuning in GNisi. Fine-tuning of the two codes does not, at least for the data we present here, improve the results at the cost of vastly increased runtime. As the coniii package contains differing sampling methods for solving the inverse Ising method, we choose a distinct method  from ACE, the Monte Carlo histogram method~\citep{broderick2007faster}.

In Fig.~\ref{fig:ising_matrix_test} we show  reconstructed Ising models for the system using the differing methods, along with the ground truth matrix in panel (b).
\begin{figure}[h!]
  \centering
  \begin{subfigure}[h]{0.3\textwidth}
    \centering
    \includegraphics[width=0.9\textwidth]{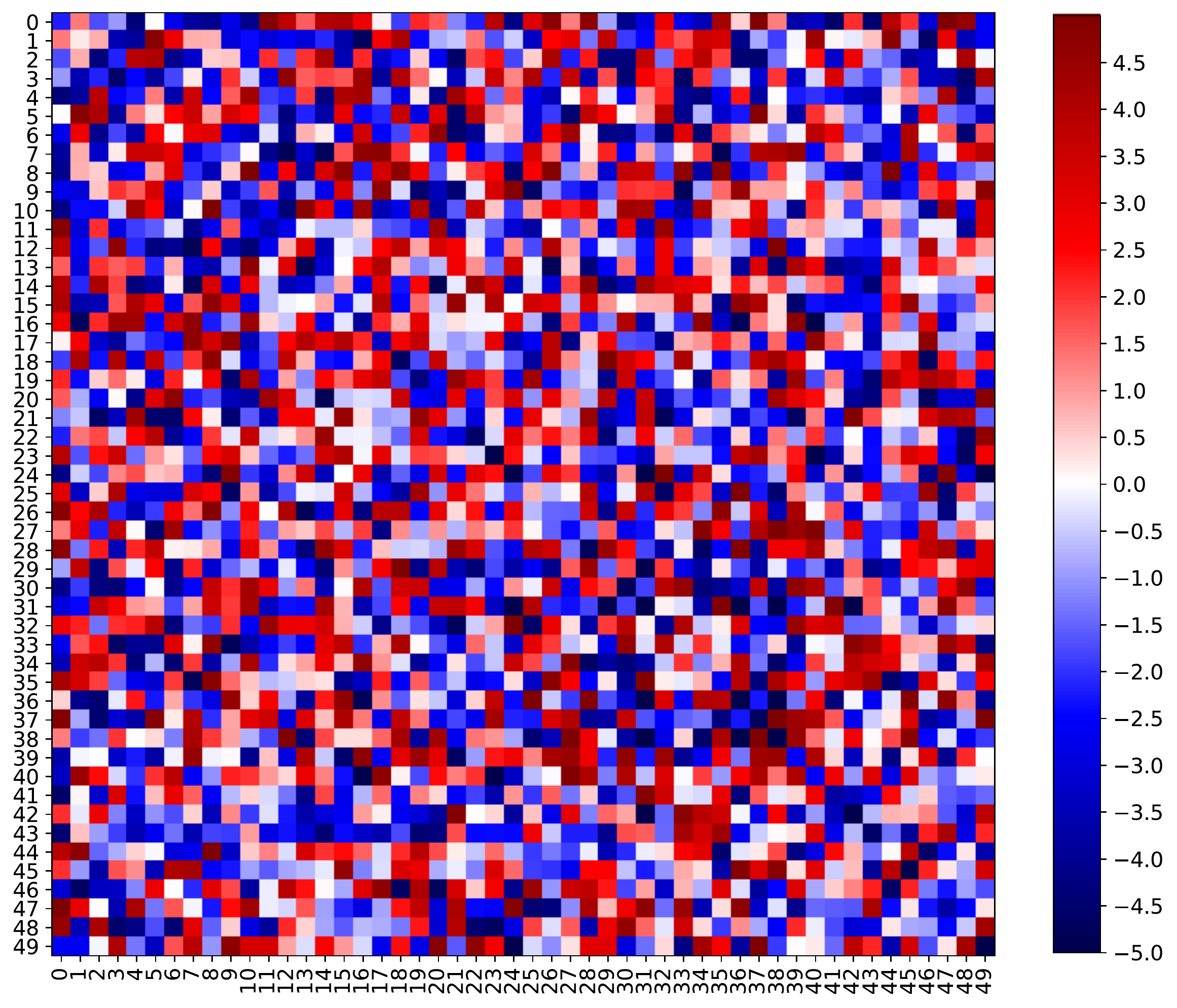}
    \caption{GNisi}
  \end{subfigure}
  \begin{subfigure}[h]{0.3\textwidth}
    \centering
    \includegraphics[width=0.9\textwidth]{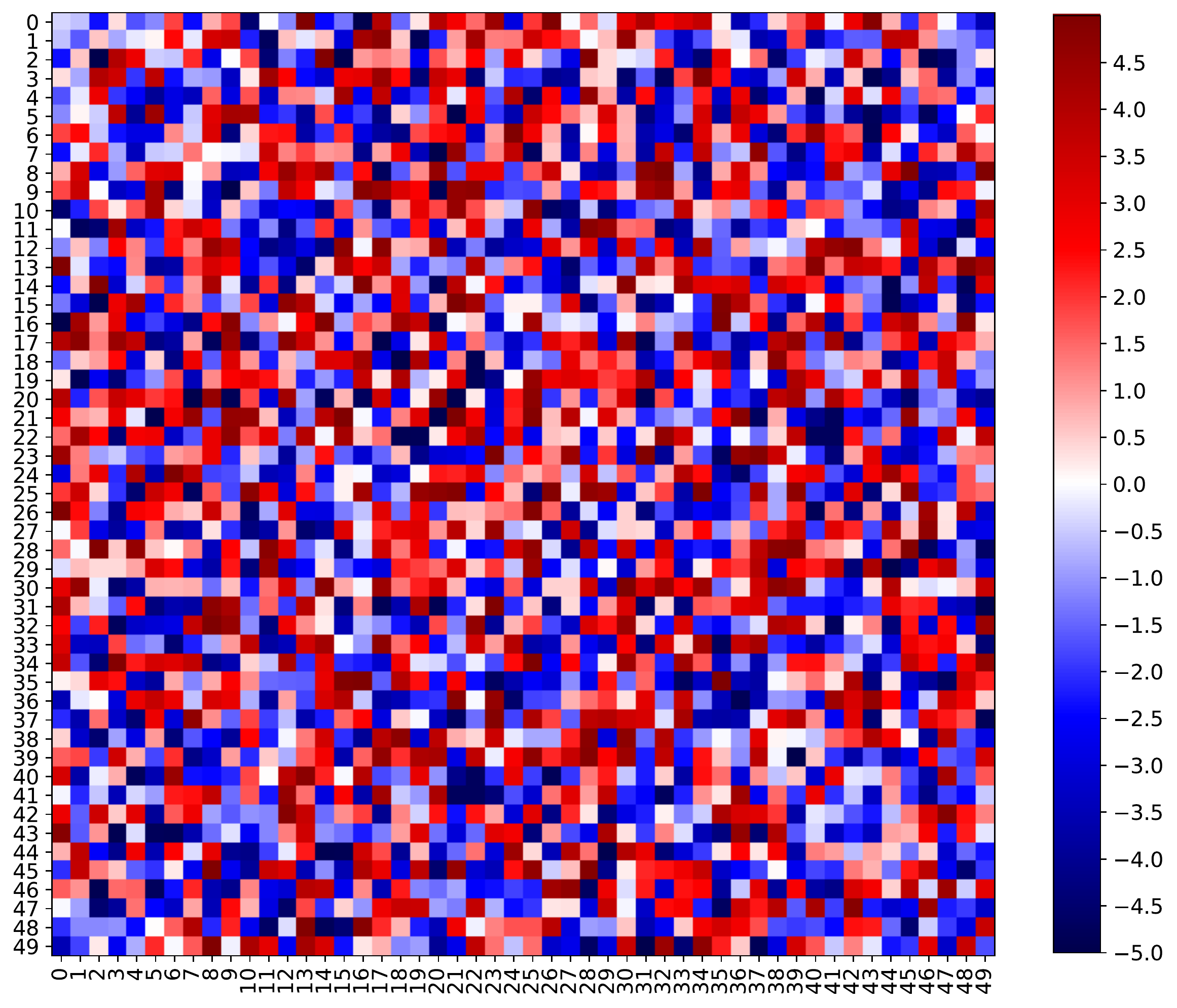}
    \caption{Ground truth}
  \end{subfigure}
   \begin{subfigure}[h]{0.3\textwidth}
    \centering
    \includegraphics[width=0.9\textwidth]{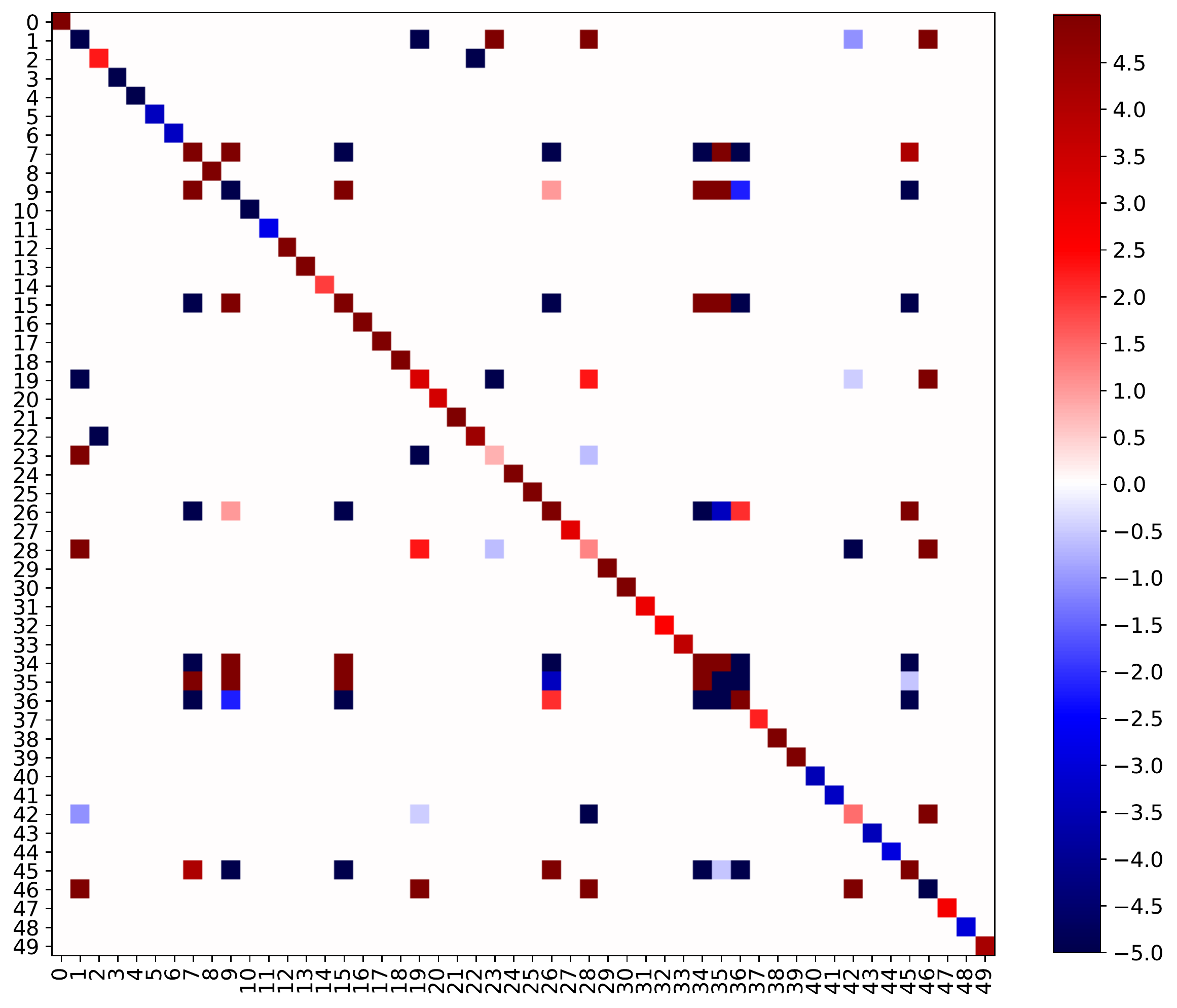}
    \caption{ACE}
      \end{subfigure}
         \begin{subfigure}[h]{0.3\textwidth}
    \centering
    \includegraphics[width=0.9\textwidth]{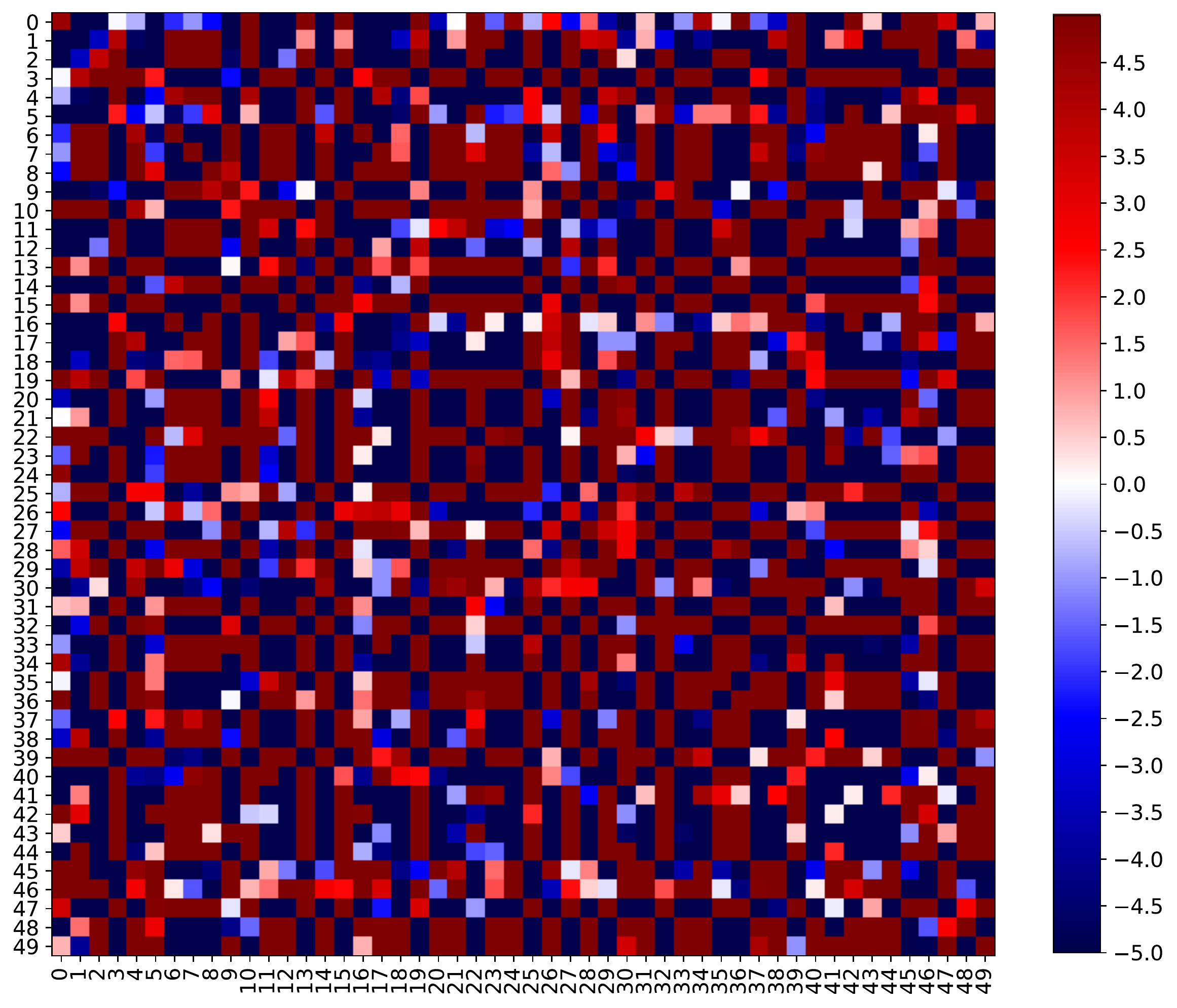}
    \caption{coniii}
      \end{subfigure}
    \caption{Reconstructed Ising models for a system of size 50 with $\beta = 10$ and 25\% sparsity in the matrix elements.} \label{fig:ising_matrix_test} 
\end{figure}
The differences in the matrices in Fig.~\ref{fig:ising_matrix_test} are hard to quantify by eye, however one can clearly tell ACE and coniii do not capture the correct overall behaviour of the system. Indeed, the mean squared error between the ground truth and the different predictions quantifies this as shown in Table~\ref{mse_params}, showing that GNisi is better than the other methods at estimating the collective behavior of the system. This is an advantage of our methodology as we do not rely on matching means and covariances; by directly using the  data, we retain much more information about the system than can be contained in the covariance matrix over the samples.

\begin{table}[h!]
\small
  \caption{MSE and Pearson correlation coefficient between the Ising parameters computed using the different methods and ground truth.}
  \label{mse_params}
  \centering
  \begin{tabular}{lll}
    \toprule
    \cmidrule(r){1-3}
    Method     &MSE & $r$    \\
    \midrule
    GNisi     &  15.38   & 0.04  \\
    ACE     & 101.53  &-0.02      \\
    coniii & 179.60& 0.02 \\
    \bottomrule
  \end{tabular}
\end{table}

It is useful to compute the Pearson correlation coefficient $r$ between the ground truth and reconstructed matrices which we show in Table~\ref{mse_params} and suggests that GNisi is not getting the individual matrix-elements correct. In Fig.~\ref{fig:hamiltonian-plot} we plot a scatter plot for the ground truth and predicted Boltzmann probability distributions, Eq.~\eqref{eq:ising_distribution}, for a sample of possible bit-strings, distinct from the bit-strings used to generate the Ising model for each method, as in general, we can evaluate the Boltzmann distribution for any possible bit-string. This can be quantified by computing the partition function, Eq.~\eqref{eq:partition_function} as well as the Pearson correlation coefficient, which we present in Table~\ref{log_z}.
 We observe that GNisi far outperforms the other methods and captures well the overall distribution. We can conclude that, despite the fact that GNisi does not reproduce individual matrix-elements, it matches the Boltzmann distribution of the ground truth very well. Therefore,  it \emph{does} produce a valid Ising model for the data it is given and we see that GNisi produces the most accurate Ising model compared to the other methods.
\begin{table}[h!]
\small
  \caption{Partition functions and correlation coefficients computed for an Ising model with differing methods.}
  \label{log_z}
  \centering
  \begin{tabular}{lll}
    \toprule
    \cmidrule(r){1-3}
    Method     & $\log(Z)$ & $r$     \\
    \midrule
    Ground truth &  55.37  & $-$    \\
    GNisi     & 52.39   &$0.44$  \\
    ACE     & 70.52      &$-0.35$  \\
    coniii & 271.41&$-0.07$ \\
    \bottomrule
  \end{tabular}
\end{table}

\begin{figure}[h!]
  \centering
  \begin{subfigure}[h]{0.3\textwidth}
    \centering
    \includegraphics[width=0.9\textwidth]{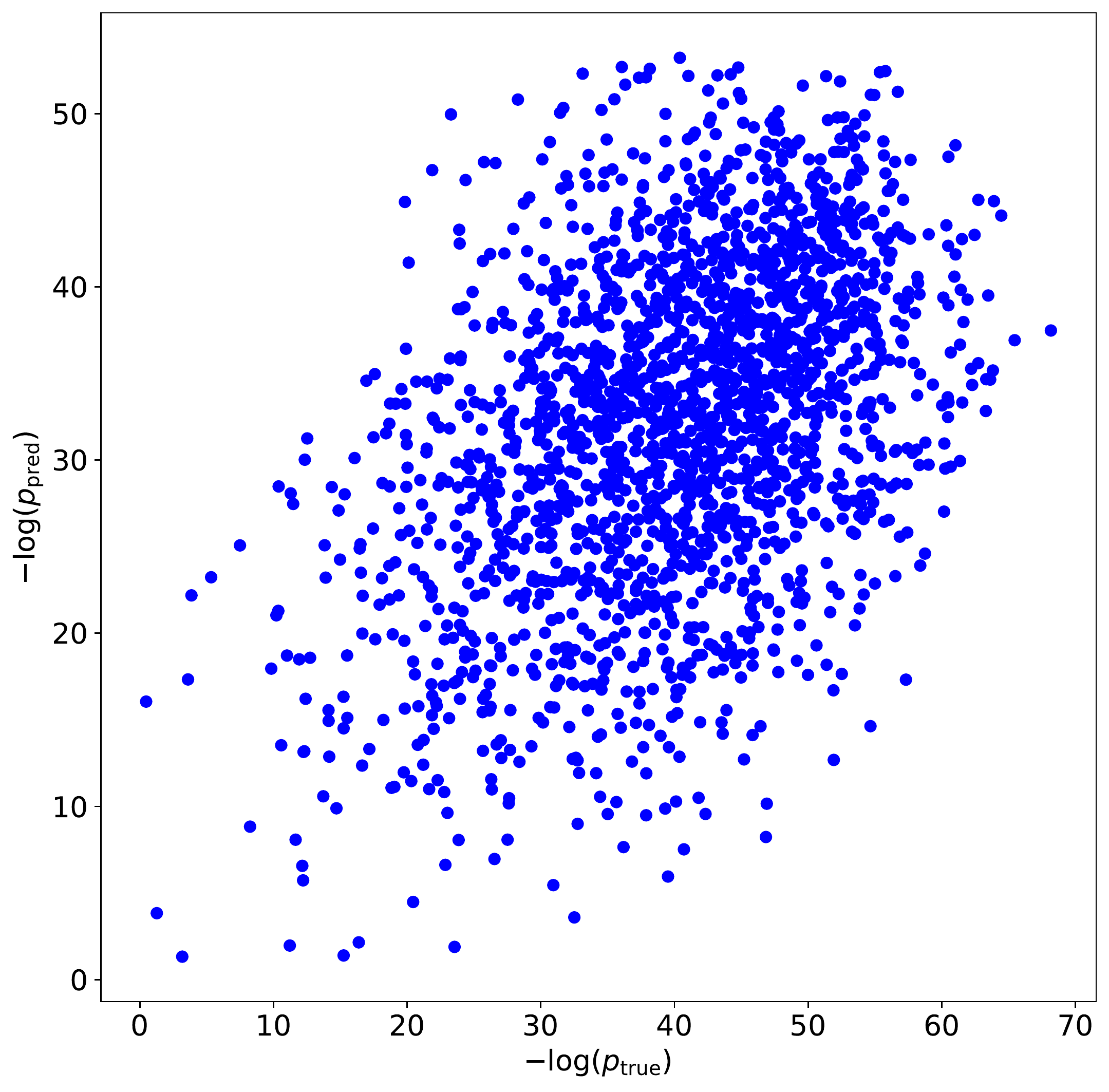}
    \caption{GNisi}
  \end{subfigure}
    \begin{subfigure}[h]{0.3\textwidth}
    \centering
    \includegraphics[width=0.9\textwidth]{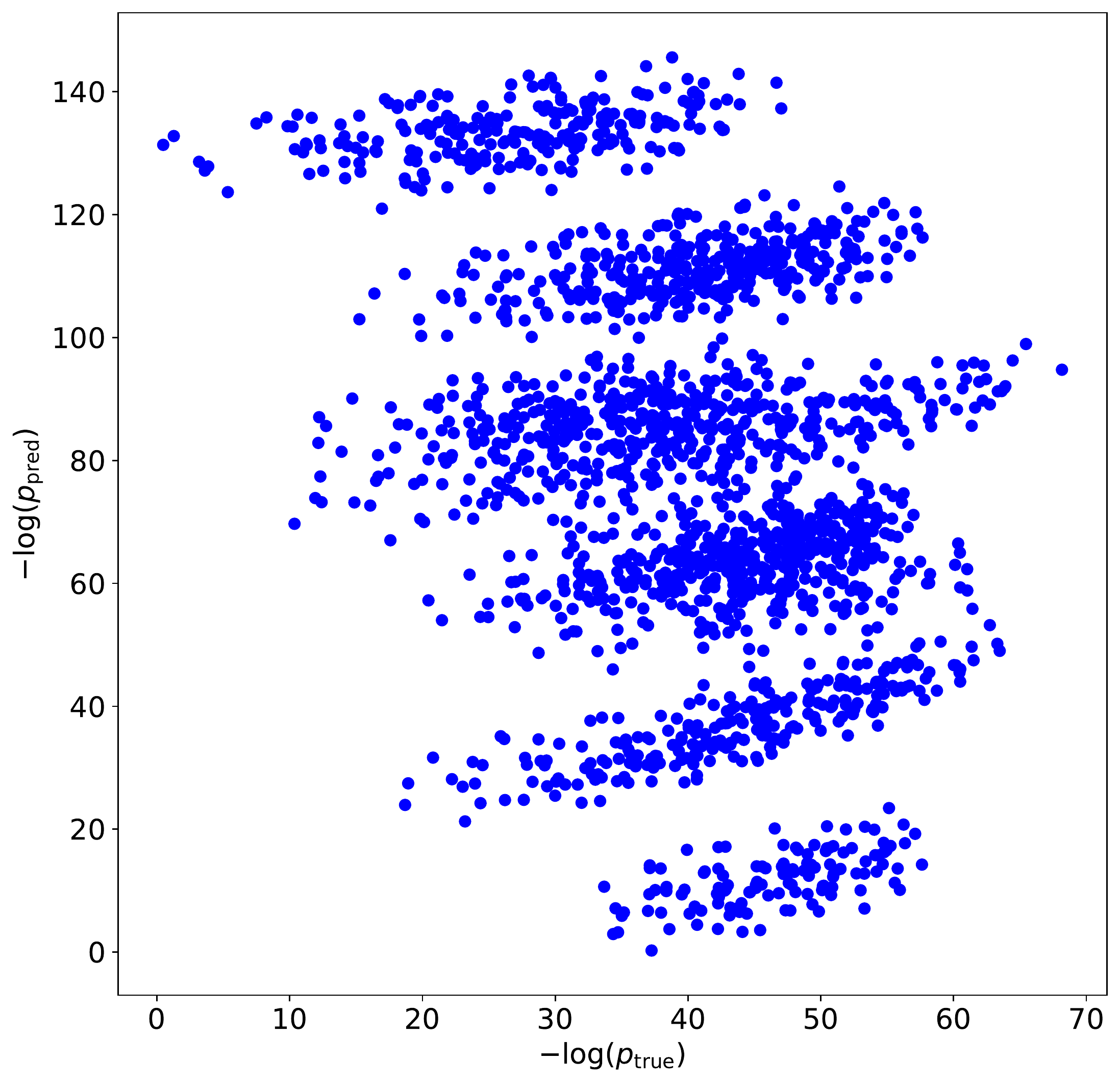}
    \caption{ACE}
  \end{subfigure}
      \begin{subfigure}[h]{0.3\textwidth}
    \centering
    \includegraphics[width=0.9\textwidth]{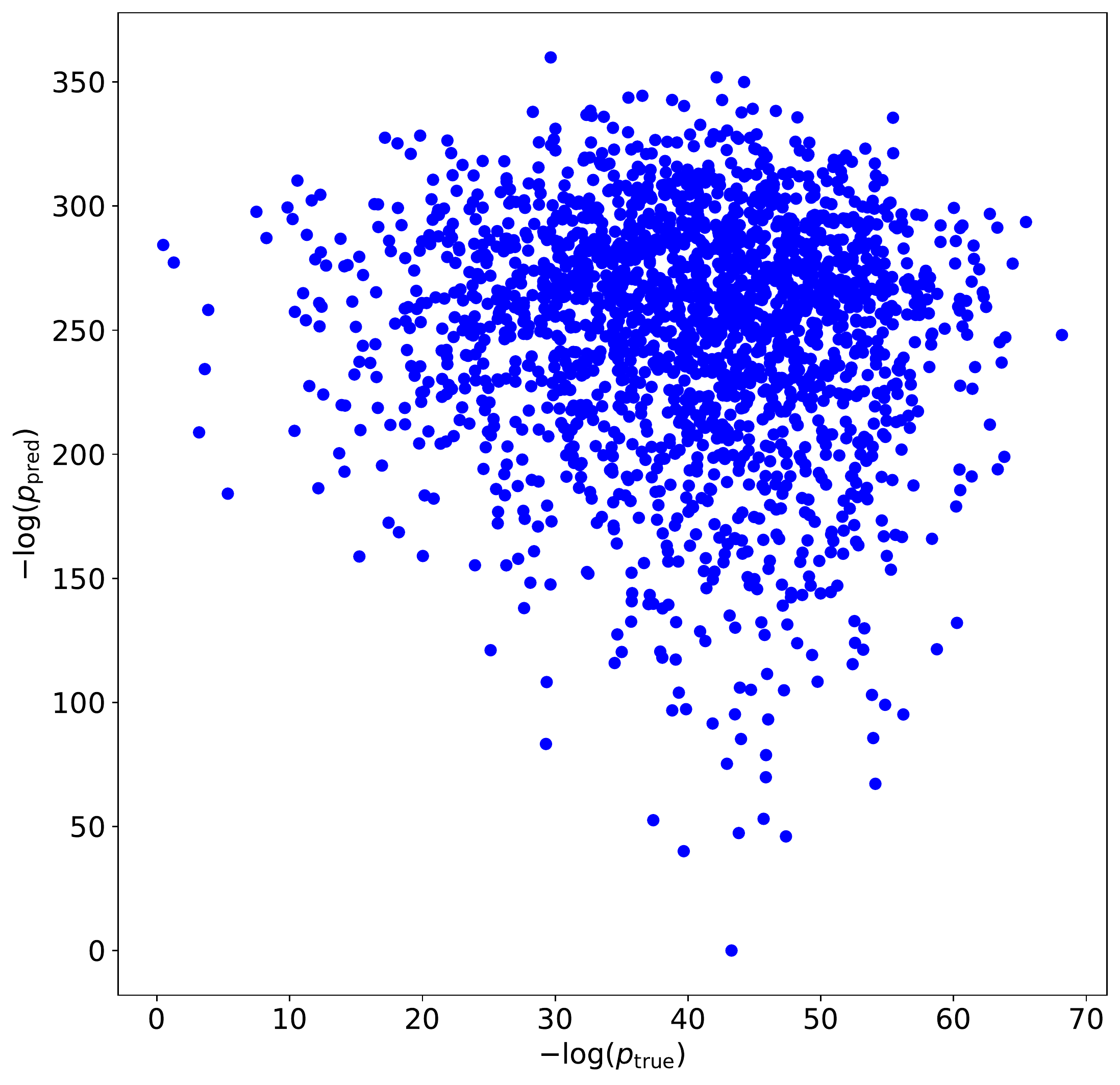}
    \caption{coniii}
  \end{subfigure}
    \caption{Scatter plots of probability distributions, Eq.~\eqref{eq:ising_distribution} computed using the different methods for a sample of possible bit-strings.} \label{fig:hamiltonian-plot}
\end{figure}

A final point of comparison is comparing the first, second and third connected moments, $m_n$, of $\xx$ (Eqs.~\eqref{eq:first_moment}, \eqref{eq:second_moment} and \eqref{eq:third_moment} respectively) and comparing to ground truth. ACE and coniii are constructed such that they produce an Ising model which matches the first and second moments, however GNisi does not have this requirement imposed in the architecture, and it is therefore a non-trivial check of our methodology that we can do so. In addition, we consider the third moment of $\xx$ as a method to more powerfully compare our methodology with the ground truth.
For the second moments, for an $n$-dimensional lattice, there are ${n \choose 2}$ possible combinations, which is not prohibitive to compute for our system. At third order, however, we have to consider ${n \choose 3}$, and we instead will consider only a sample of the full distribution. %
In Table~\ref{mse-moments-table} we show the mean squared error between the moments computed using the ground truth, and with the various methods.  In all cases we show the $n$th connected moment with either all bit-strings being 1 ($m_n(1)$), or 0 ($m_n(0)$). Unsurprisingly, both ACE and coniii match the moments very well, as they construct Ising models under the condition that the maximum entropy condition is met. It is therefore  impressive that GNisi, which does not attempt to match moments, also matches the ground truth to very high precision. We show in the Supplementary Material the results obtained using a random matrix with the same size couplings as GNisi, to explicitly show that GNisi is correctly learning features of the Ising model. Interestingly,  ACE and coniii match extremely well (to the precision shown here), suggesting the maximum entropy approach strongly constrains higher order moments.

In order to explain the behavior we observe, namely that all the methods used in this paper can match the moments of the data distribution well but, to a greater or lesser extent, not match the individual parameters of the ground truth Ising model, we note that we have a sloppy model~\citep{BrownS03,Brown_2004}. A sloppy model is one which is poorly constrained as there exist a very large number of parameter choices which can match the data distribution well. As we need to match $n^2$ parameters for an Ising model of size $n$, this behaviour can be expected to worsen as we increase the size of the system. It is therefore not surprising that the 3 different methods match the moments of the ground truth matrix with 3 distinct predictions for the Ising model. It is for this reason we argue that the ability to correctly reconstruct the Boltzmann distribution is the fundamental test for whether a generated model is accurate; in Fig.~\ref{fig:hamiltonian-plot} we observe that the predictions from GNisi correlates with the ground truth bit-string by bit-string, which strongly suggests that GNisi has correctly learned underlying features about the Ising model. 

\begin{table}
\small
  \caption{Mean squared error between the $n$th moment computed using ground truth and the differing methods. }
  \label{mse-moments-table}
  \centering
  \begin{tabular}{llll}
    \toprule
        \multicolumn{3}{r}{MSE}                   \\
    \cmidrule(r){2-4}
    Moment     & GNisi & ACE  & coniii      \\
    \midrule
    $m_1(1)$     & 0.03   & 0.11 &0.11  \\
     $m_2(1)$   & $1.60\cdot 10^{-5}$ &     $3.27\cdot 10^{-7}$ & $ 3.27\cdot 10^{-7}$ \\
          $m_2(0)$   &$1.60\cdot 10^{-5}$ &   $3.27\cdot 10^{-7}$    &$3.27\cdot 10^{-7}$ \\
          $m_3 (0)$   &$ 6.33\cdot 10^{-5}$ & $6.83\cdot 10^{-7}$  & $6.83\cdot 10^{-7}$ \\
               $m_3 (1)$  & 0.0  &0.0  &0.0\\
    \bottomrule
  \end{tabular}
\end{table}

\subsection{Results on the CCLE dataset}
In this section we apply the pre-trained GNisi model to two distinct datasets of genes from the CCLE dataset, each of which  are known to be highly mutually correlated in order to provide an estimate of the ground truth. We binarize the continuous data by identifying the $q$-level quantile, and setting the gene expression to 0 below the $q$-th quantile, and 1 above it. In order to prevent the binarized data from being highly skewed, we set $q=0.25$ throughout. As above, where we compare GNisi and ACE against ground truth, we will show results run using ACE default settings. Due to the poor quality of the results with coniii on the synthetic data, we will not include a comparison with the package here.

\subsubsection{Highly correlated genes}
\begin{figure}[h!]
  \centering
  \begin{subfigure}[h]{0.35\textwidth}
    \centering
    \includegraphics[width=0.9\textwidth]{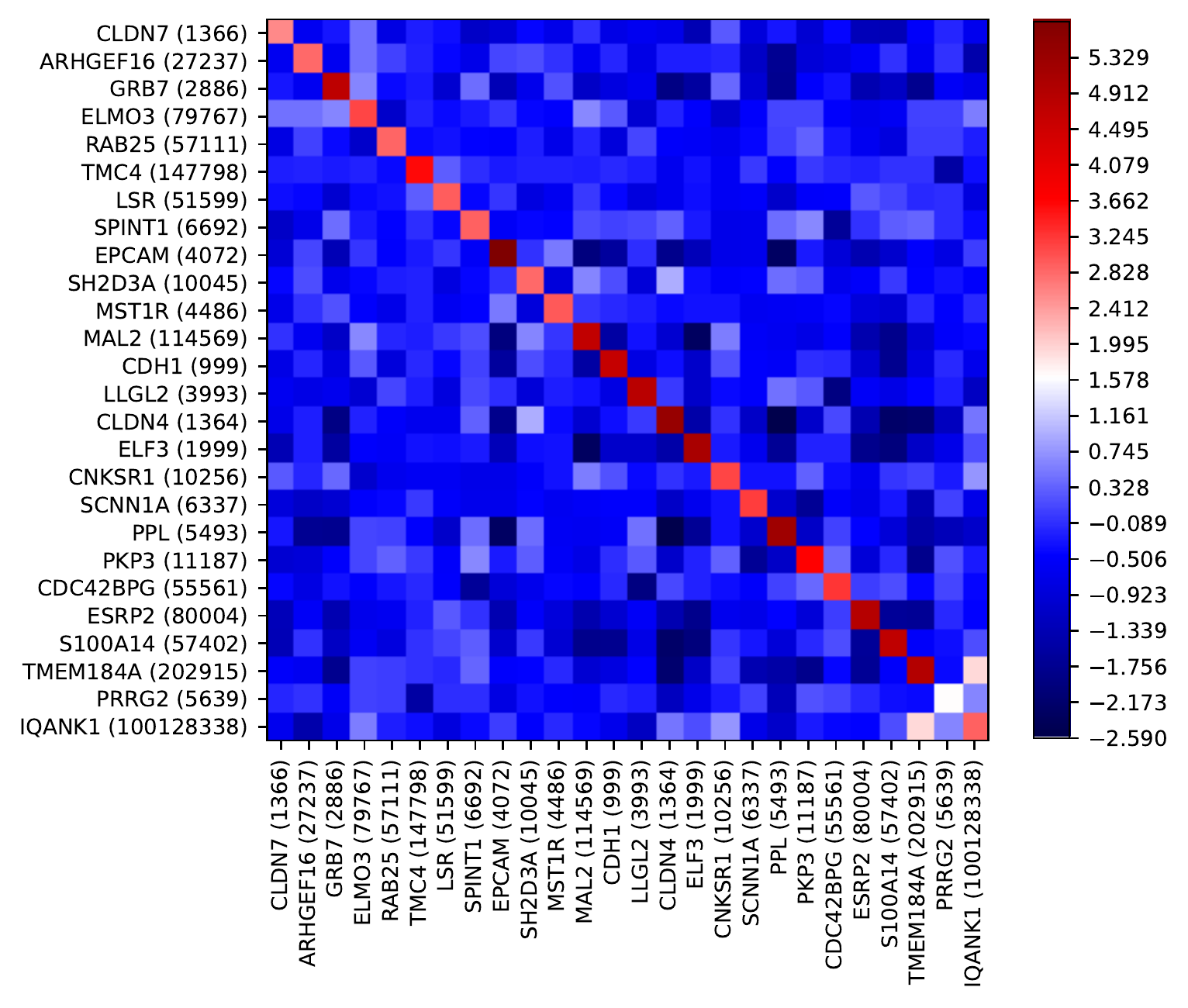}
    \caption{GNisi}
  \end{subfigure}
  \begin{subfigure}[h]{0.35\textwidth}
    \centering
    \includegraphics[width=0.9\textwidth]{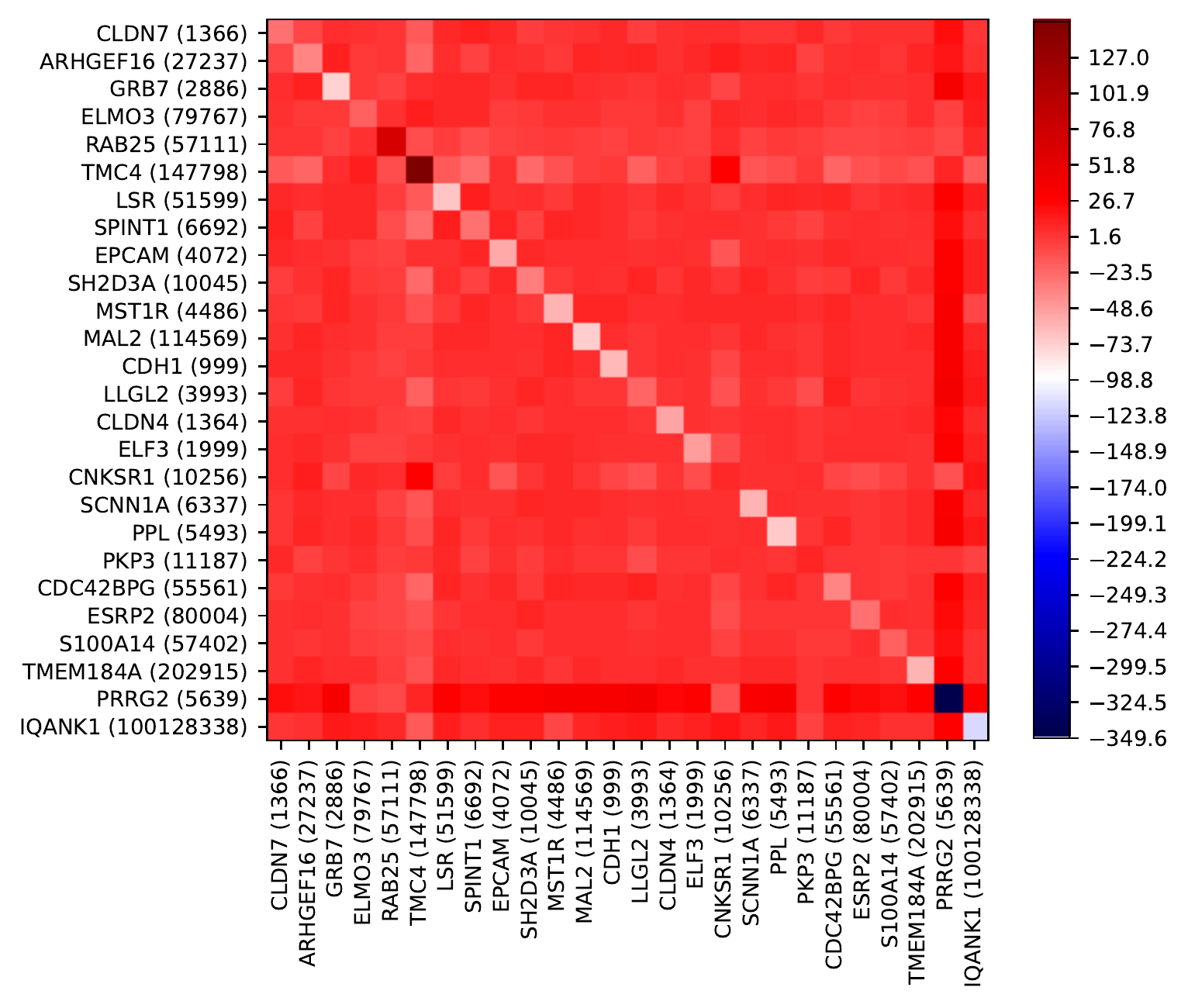}
    \caption{ACE}
      \end{subfigure}
    \caption{Reconstructed Ising models for a group of genes from the CCLE dataset known to be highly mutually correlated.} \label{fig:ccle_matrix} 
\end{figure}

In Fig.~\ref{fig:ccle_matrix} we show  reconstructed Ising models for the first system using the two differing methods. We can see that both GNisi and ACE correctly produce a highly correlated matrix, however the couplings produced by ACE are an order of magnitude larger than those produced by GNisi. %

\begin{figure}[h!]
  \centering
  \begin{subfigure}[h]{0.7\textwidth}
    \centering
    \includegraphics[width=0.8\textwidth]{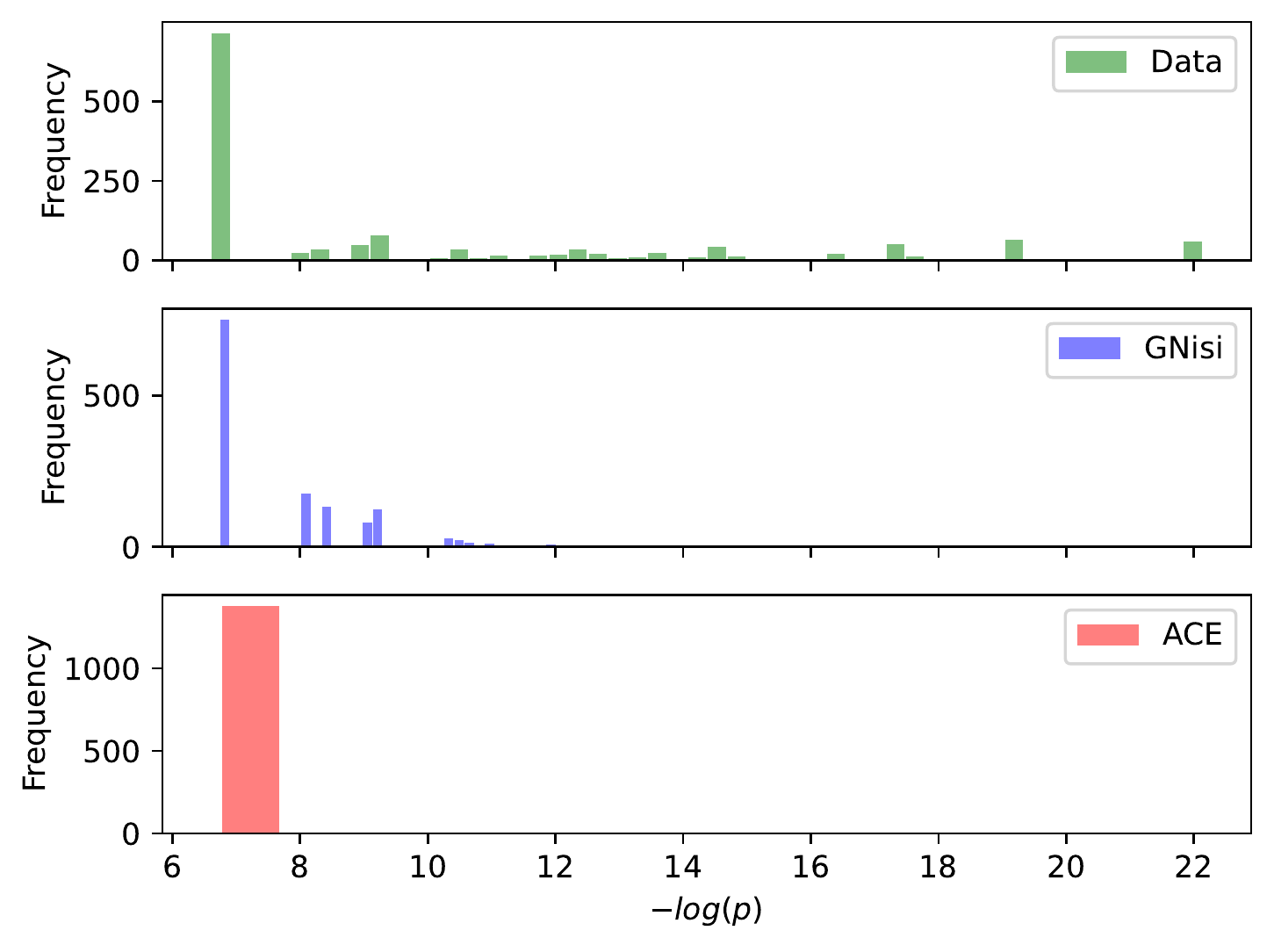}
  \end{subfigure}
    \caption{Log Boltzmann probability distribution for the different methods using CCLE data.} \label{fig:reconstructed_boltzmann} 
\end{figure}

We argued in the previous section that the fundamental test to determine the accuracy of our reconstructed Ising model is to compare the Boltzmann distribution constructed from the Ising model against the data. Unlike in Sec.~\ref{sec:unseen}, where we compare samples of the partition function against ground truth in Fig.~\ref{fig:hamiltonian-plot}, in this example (and all real-world applications) we do not know the true Ising model which describes the data. In Fig.~\ref{fig:reconstructed_boltzmann} we show a sample of the log Boltzmann probability distributions computed using the observed data samples and the reconstructed Ising models from GNisi and ACE. As there are $N=1376$ samples in the observed dataset, we sample $N$ possible bit-strings from the total number of combinations and compute the probability of each permutation, given the underlying Ising model for GNisi and ACE. It is clear that, while GNisi does not perfectly match the observed distribution, as the distribution has a smaller tail, it matches very closely. On the other hand, ACE is strongly peaked at $-\log(p) \sim 7$, which is correct, but has no other features.

We now turn to the first, second and third moments predicted by the different models. In Table~\ref{mse-moments-table-ccle} we show mean squared error between the sampled moments and the observed moments from the data distribution. We see that on the whole, both methods agree well with the data, with GNisi having a larger errors for most moments. Interestingly, for this subset of genes, we find that ACE outperforms GNisi with respect to matching moments as shown in Table~\ref{mse-moments-table-ccle}. It should be noted, however, that matching the low-order moments of the training data does not necessarily capture the data distribution better by all measures, and there was no ground truth of a known underlying Ising model to compare to in this case of real experimental data.  Indeed, in this case it is clear that GNisi produces the more accurate Ising model when judged by how close the predicted log Boltzmann probability distribution is to the observed distribution. 
\begin{table}
\small
  \caption{Mean squared error between the $n$th moment computed using ground truth and ACE for the CCLE data. }
  \label{mse-moments-table-ccle}
  \centering
  \begin{tabular}{lll}
    \toprule
        \multicolumn{2}{r}{MSE}                   \\
    \cmidrule(r){2-3}
    Moment     & GNisi & ACE        \\
    \midrule
    $m_1(1)$     &  0.36  &   0.43\\
     $m_2(1)$   & $4.15 \cdot 10^{-4}$ &$1.58 \cdot 10^{-6}  $     \\
          $m_2(0)$   &$4.17\cdot 10^{-4}  $&$1.71 \cdot 10^{-6}  $ \\
          $m_3 (1)$   &$3.29\cdot 10^{-4} $& $3.29\cdot 10^{-6}$ \\
               $m_3 (0)$  &  $1.87 $ & $9.13\cdot 10^{-4}$ \\
    \bottomrule
  \end{tabular}
\end{table}
%MSE for means (data, pred):  0.35968728588261933
%MSE for pairwise (1,1):  0.004154142676290809
%MSE for pairwise (0,0):  0.004170339685689544
%MSE for 3rd moment (1,1,1):  0.0003287377105626979
%MSE for 3rd moment (0,0,0):  1.8672985920139722
%MSE for means (ground truth, ACE):  0.42556634705855445
%MSE for pairwise (0,0):  1.707996778242788e-06
%MSE for pairwise (1,1):  1.5770112879688031e-06
%MSE for 3rd moment (1,1,1):  0.0003287377105626979
%MSE for 3rd moment (0,0,0):  9.128486347715579e-06

\subsubsection{Genes with known mutual relation to each other and to BRCA1 }
\begin{figure}[h!]
  \centering
  \begin{subfigure}[h]{.35\textwidth}
    \centering
    \includegraphics[width=0.9\textwidth]{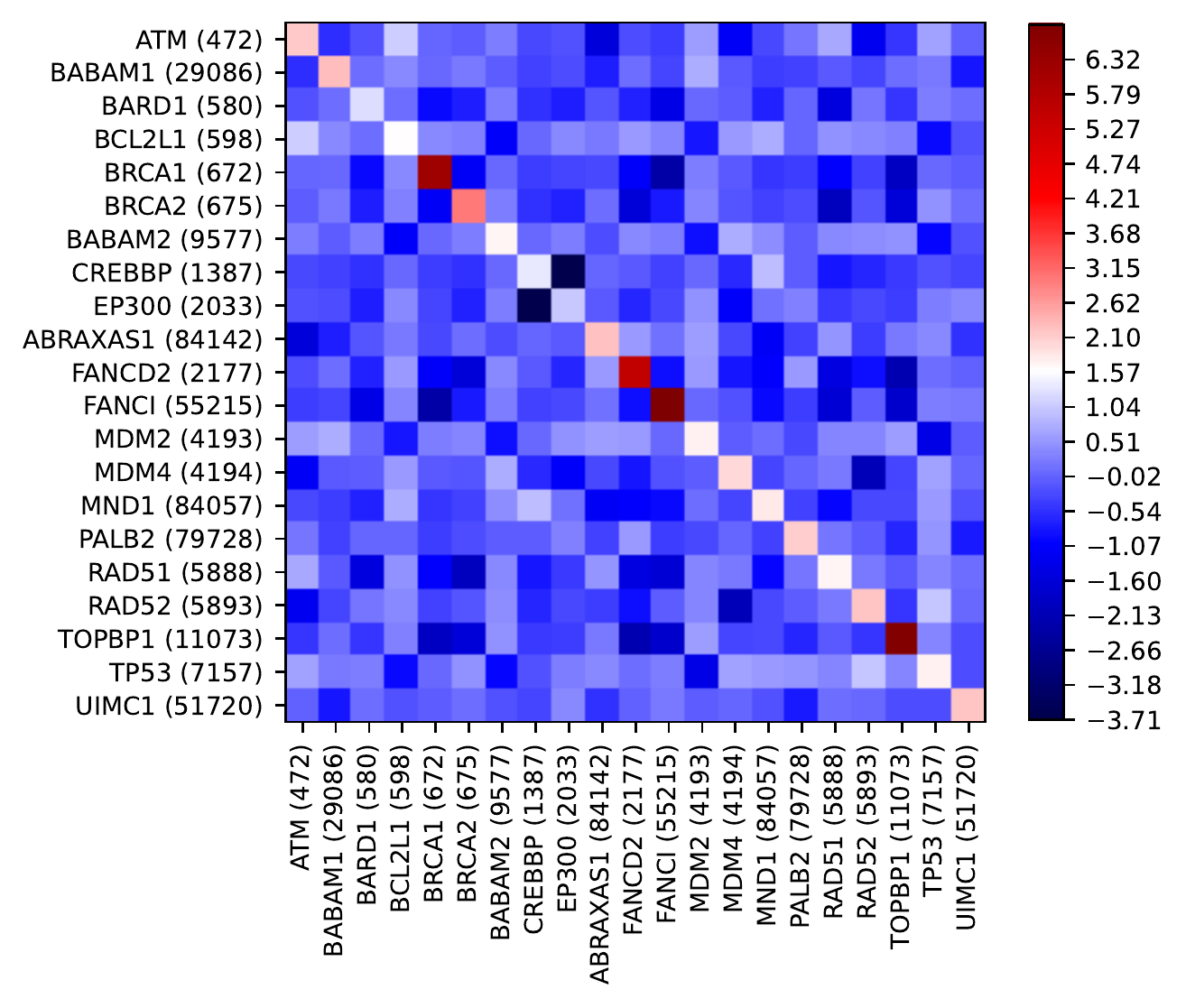}
    \caption{GNisi}
  \end{subfigure}
  \begin{subfigure}[h]{0.35\textwidth}
    \centering
    \includegraphics[width=0.9\textwidth]{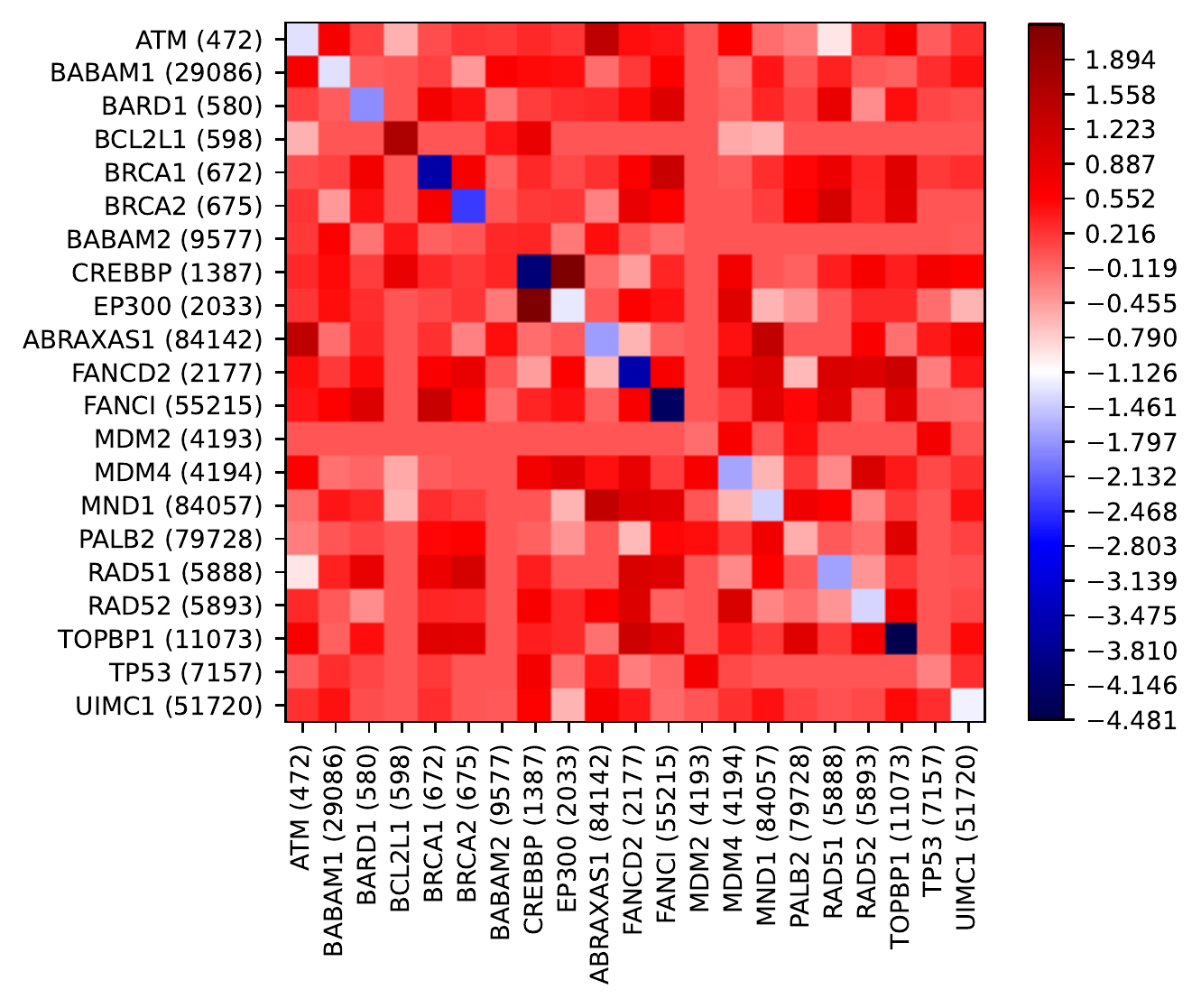}
    \caption{ACE}
      \end{subfigure}
    \caption{Reconstructed Ising models for a group of genes from the CCLE dataset known to be highly mutually correlated with BRCA1.} \label{fig:ccle_matrix_brca1} 
\end{figure}

Finally, we turn to the second distinct set of genes. As before, some biological ground truth is known about the 21 genes in this dataset; here we pick the subset of genes with known functional relation to BRCA1 and to each other. In Fig.~\ref{fig:ccle_matrix_brca1} we show the reconstructed Ising models using GNisi and ACE; as before we see that both methods find a high level of mutual correlation in the matrix, however GNisi finds much stronger couplings than ACE.

\begin{figure}[h!]
  \centering
  \begin{subfigure}[h]{0.7\textwidth}
    \centering
    \includegraphics[width=0.8\textwidth]{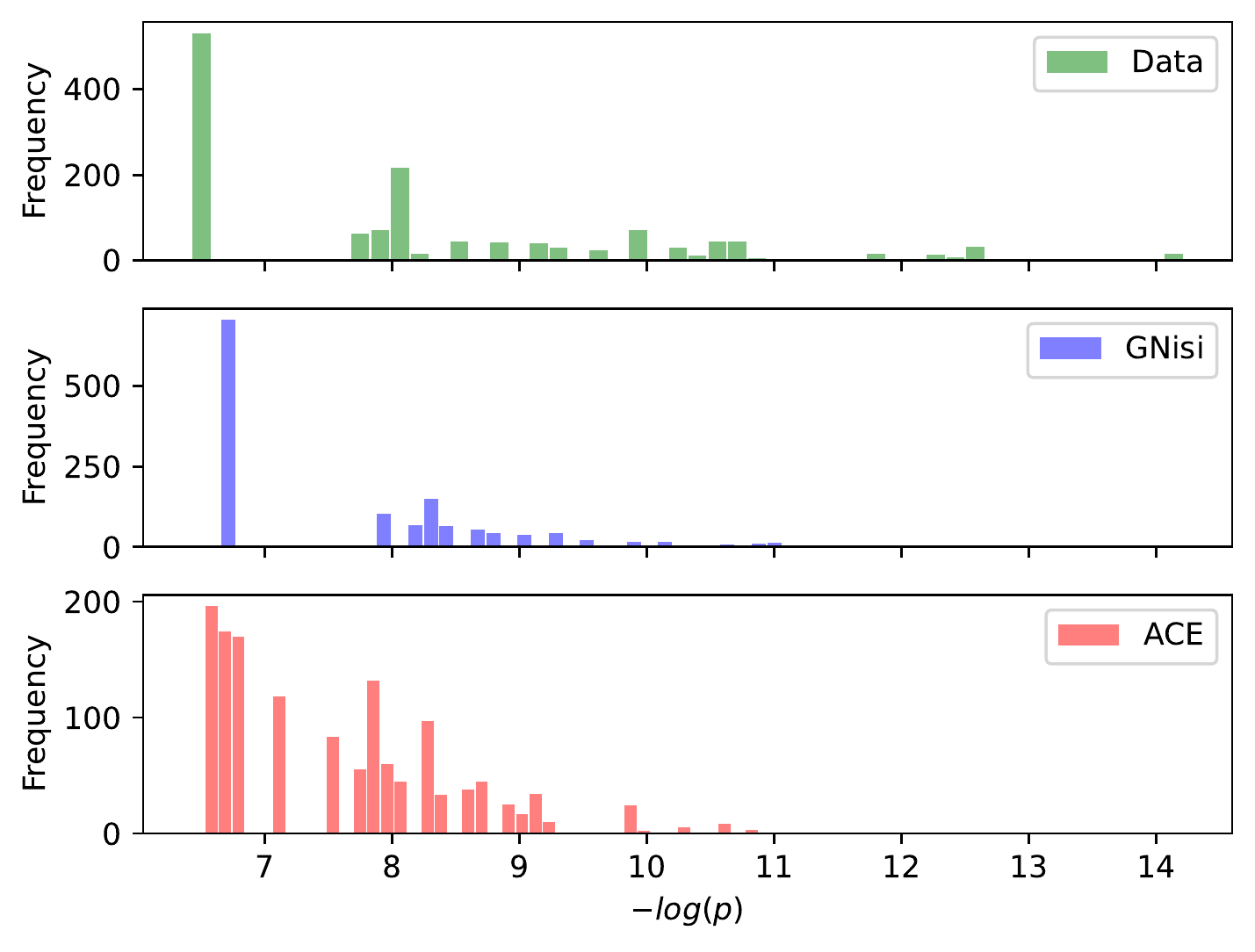}
  \end{subfigure}
    \caption{Log Boltzmann probability distribution for the different methods using CCLE data for genes  known to be highly mutually correlated with BRCA1.} \label{fig:reconstructed_boltzmann_brca1} 
\end{figure}

 To again determine the accuracy of the reconstructions, we plot the observed and sampled log Boltzmann probability distributions in Fig.~\ref{fig:reconstructed_boltzmann_brca1}. As with the previous dataset, GNisi matches the log Boltzmann probability distribution for the data  better than ACE, although ACE performs much better with this dataset than the previous one. We again see that GNisi matches the shape of the distribution somewhat better than ACE. We show the MSE between moments for this dataset in Table~\ref{mse-moments-table-ccle-brca1}; again both GNisi and ACE perform well, with ACE somewhat worse than GNisi in this case.

\begin{table}
\small
  \caption{Mean squared error between the $n$th moment computed using ground truth and ACE for the CCLE data  for genes  known to be highly mutually correlated with BRCA1. }
  \label{mse-moments-table-ccle-brca1}
  \centering
  \begin{tabular}{lll}
    \toprule
        \multicolumn{2}{r}{MSE}                   \\
    \cmidrule(r){2-3}
    Moment     & GNisi & ACE        \\
    \midrule
    $m_1(1)$     &0.43   & 0.27  \\
     $m_2(1)$   &  $7.51 \cdot 10^{-5}$&  $5.16\cdot 10^{-4}$     \\
          $m_2(0)$   &$7.55\cdot 10^{-5}$  &$5.18\cdot 10^{-4}$     \\
          $m_3 (1)$   & $2.50\cdot 10^{-4}$ &  $2.50\cdot 10^{-4}$ \\
               $m_3 (0)$  &   $2.89\cdot 10^{-5}$  &$0.027$  \\
    \bottomrule
  \end{tabular}
\end{table}
%
%MSE for means (data, pred):  0.4333769477884398
%MSE for pairwise (1,1):  7.506501415980895e-05
%MSE for pairwise (0,0):  7.43836959879702e-05
%MSE for 3rd moment (1,1,1):  0.00024995484422232143
%MSE for 3rd moment (0,0,0):  0.0002889692614153833
%MSE for means (ground truth, ACE):  0.2717267908303461
%MSE for pairwise (0,0):  0.0005178584416850922
%MSE for pairwise (1,1):  0.0005155693902654067
%MSE for 3rd moment (1,1,1):  0.00024995484422232143
%MSE for 3rd moment (0,0,0):  0.027068714847028368

\section{Conclusion}
In this paper we have presented a novel method for solving the inverse Ising problem. Inverse Ising methods have been successfully applied to a wide range of biological systems, however the computational difficulty of solving the inverse problem often prevents their use, or necessitates the use of approximate methods such as Gaussian approximations or pseudo-likelihood methods. With GNisi, we have presented a fast, accurate and easy-to-use method for solving inverse Ising models using graph neural networks. 

We have presented results where we apply GNisi to both synthetic and real data, and compared our results with the current state-of-the-art codes. We find that across the board, GNisi outperforms other methods. Additionally, GNisi requires no fine-tuning; once the data has been collected, it is sufficient to provide it to the trained model and an Ising model that accurately describes the data is rapidly produced. It is important to note that the results presented here for GNisi are with 1000 Monte Carlo samples per model; we expect that results will improve upon the generation of more data.

It should be noted that the choice of what random distribution of Ising couplings to sample in the Monte Carlo simulation data used to train GNisi should affect the kinds of Ising models it is capable of reconstructing: in an instance where the best Ising model of the data would be a highly rare event in the space of models considered in the training data, GNisi might be expected to fail at constructing a useful model, whether for protein structure or neuronal firing.

% Authors are required to include a statement of the broader impact of their work, including its ethical aspects and future societal consequences. 
% Authors should discuss both positive and negative outcomes, if any. For instance, authors should discuss a) 
% who may benefit from this research, b) who may be put at disadvantage from this research, c) what are the consequences of failure of the system, and d) whether the task/method leverages
% biases in the data. If authors believe this is not applicable to them, authors can simply state this.

% Use unnumbered first level headings for this section, which should go at the end of the paper. {\bf Note that this section does not count towards the eight pages of content that are allowed.}

% \begin{ack}
% Use unnumbered first level headings for the acknowledgments. All acknowledgments
% More information about this disclosure can be found at: \url{https://neurips.cc/Conferences/2020/PaperInformation/FundingDisclosure}.

% Do {\bf not} include this section in the anonymized submission, only in the final paper. You can use the \texttt{ack} environment provided in the style file to autmoatically hide this section in the anonymized submission.
% \end{ack}

% \def\bibfont{\small}
\bibliographystyle{plainnat}
\bibliography{biblio}

\appendix
\section{Comparison between GNisi and ACE for a high temperature system}

In the main paper we compared our results for a representative model from the test set at low temperature. Here we will present the same results, performing the same comparisons, for an Ising model from the test set at a temperature one of order magnitude higher, namely an Ising model of size 50 with 75\% sparsity in the matrix elements for $\beta = 1$. As before, we run the ACE and coniii packages using default settings, in order to prevent fine-tuning the results on a model-by-model basis. The MSE between the ground truth and differing methods is shown in Table~\ref{mse_params}.

In Fig.~\ref{fig:ising_matrix_test} we show  reconstructed Ising models for the system using the differing methods, along with the ground truth matrix in panel (b).
\begin{figure}[h!]
  \centering
  \begin{subfigure}[h]{0.45\textwidth}
    \centering
    \includegraphics[width=0.9\textwidth]{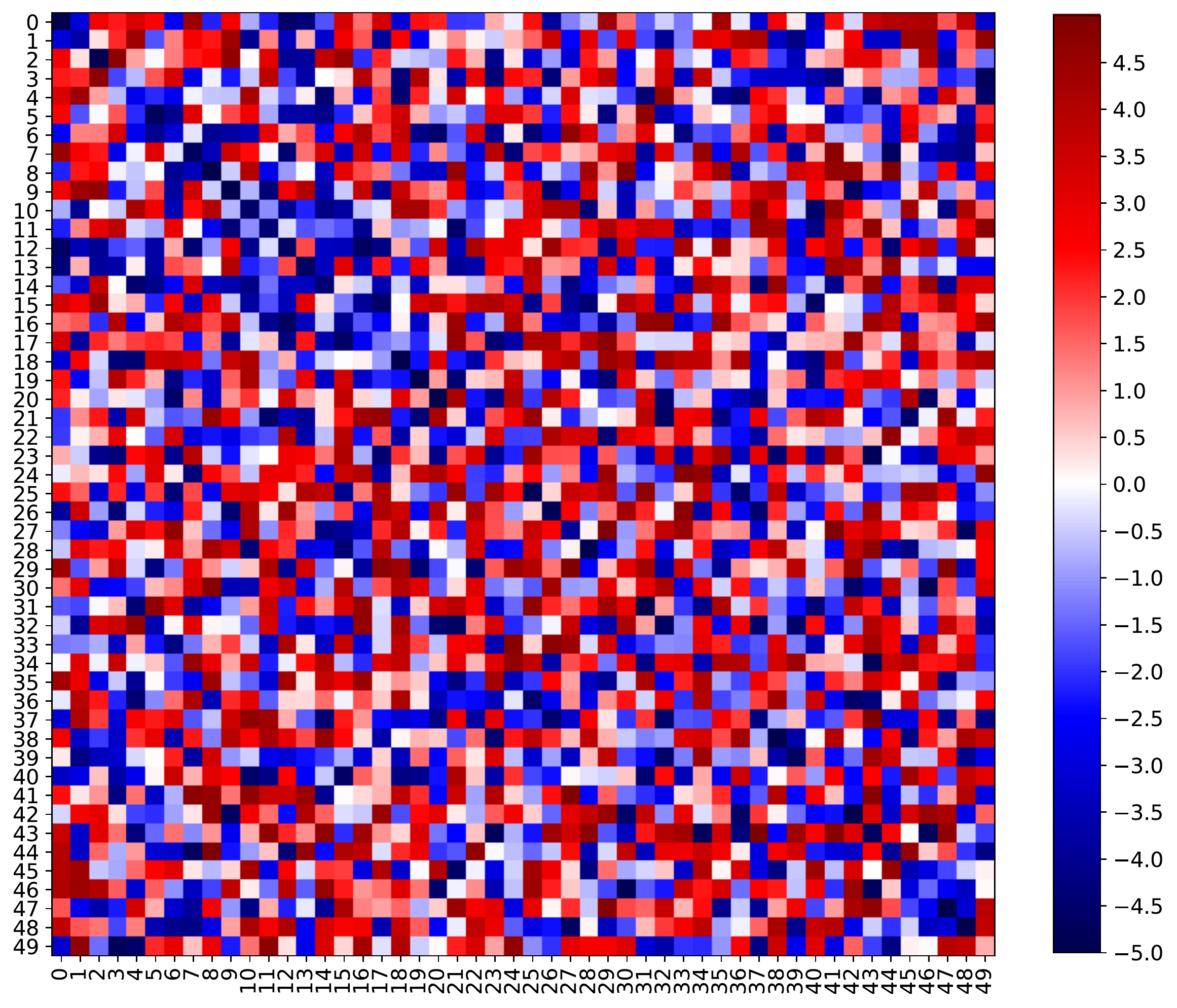}
    \caption{GNisi}
  \end{subfigure}
  \begin{subfigure}[h]{0.45\textwidth}
    \centering
    \includegraphics[width=0.9\textwidth]{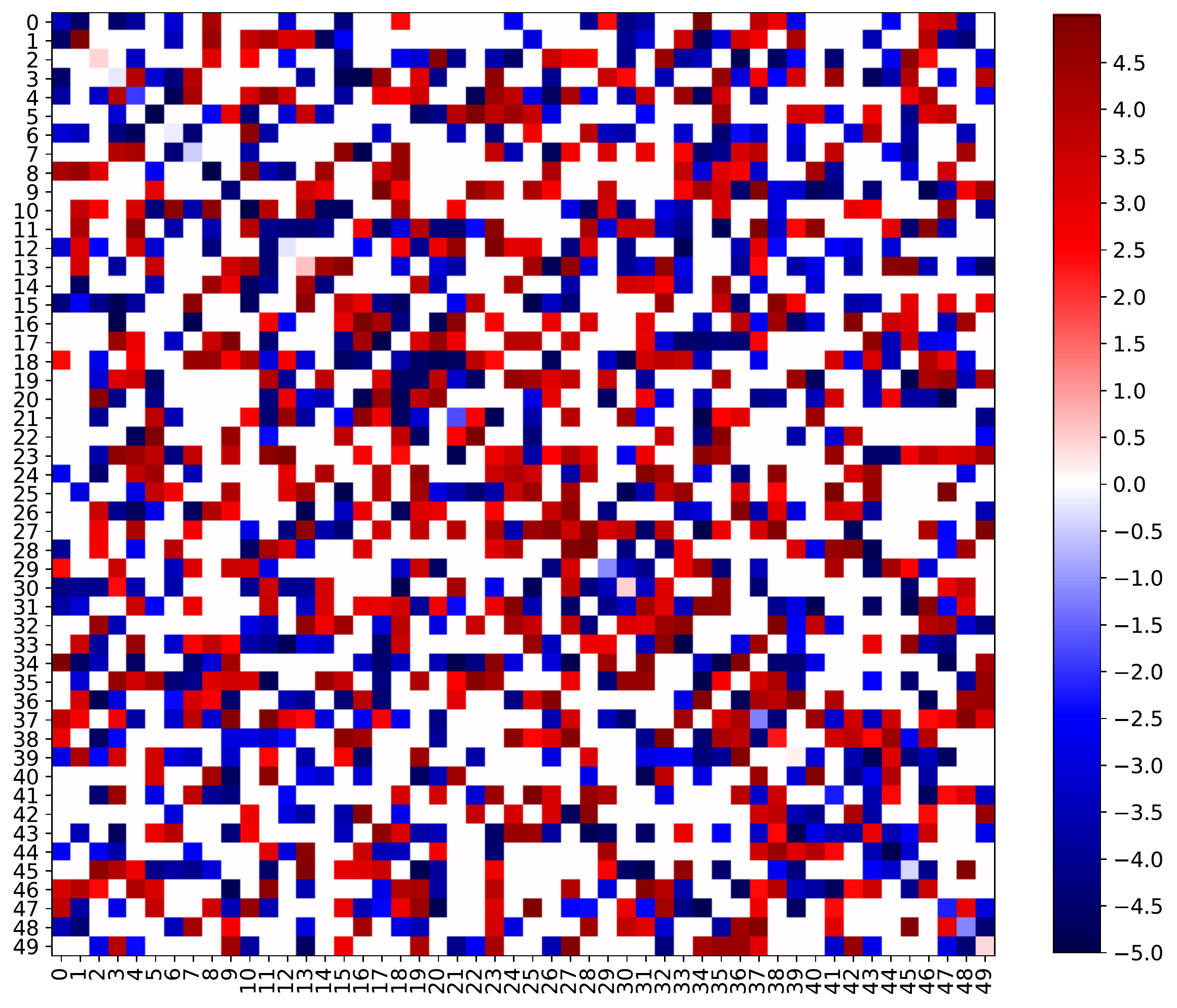}
    \caption{Ground truth}
  \end{subfigure}
   \begin{subfigure}[h]{0.45\textwidth}
    \centering
    \includegraphics[width=0.9\textwidth]{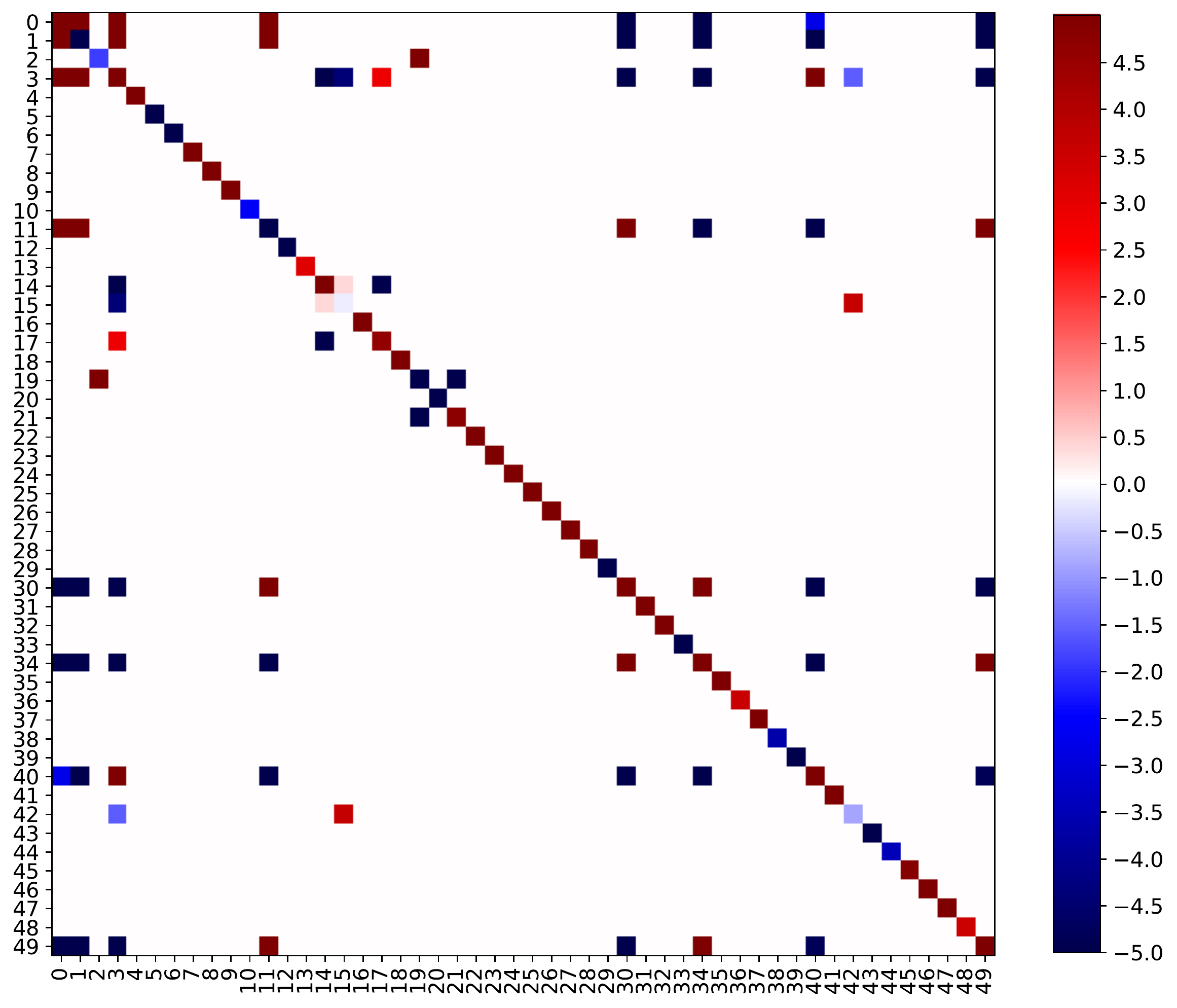}
    \caption{ACE}
      \end{subfigure}
      \begin{subfigure}[h]{0.45\textwidth}
    \centering
    \includegraphics[width=0.9\textwidth]{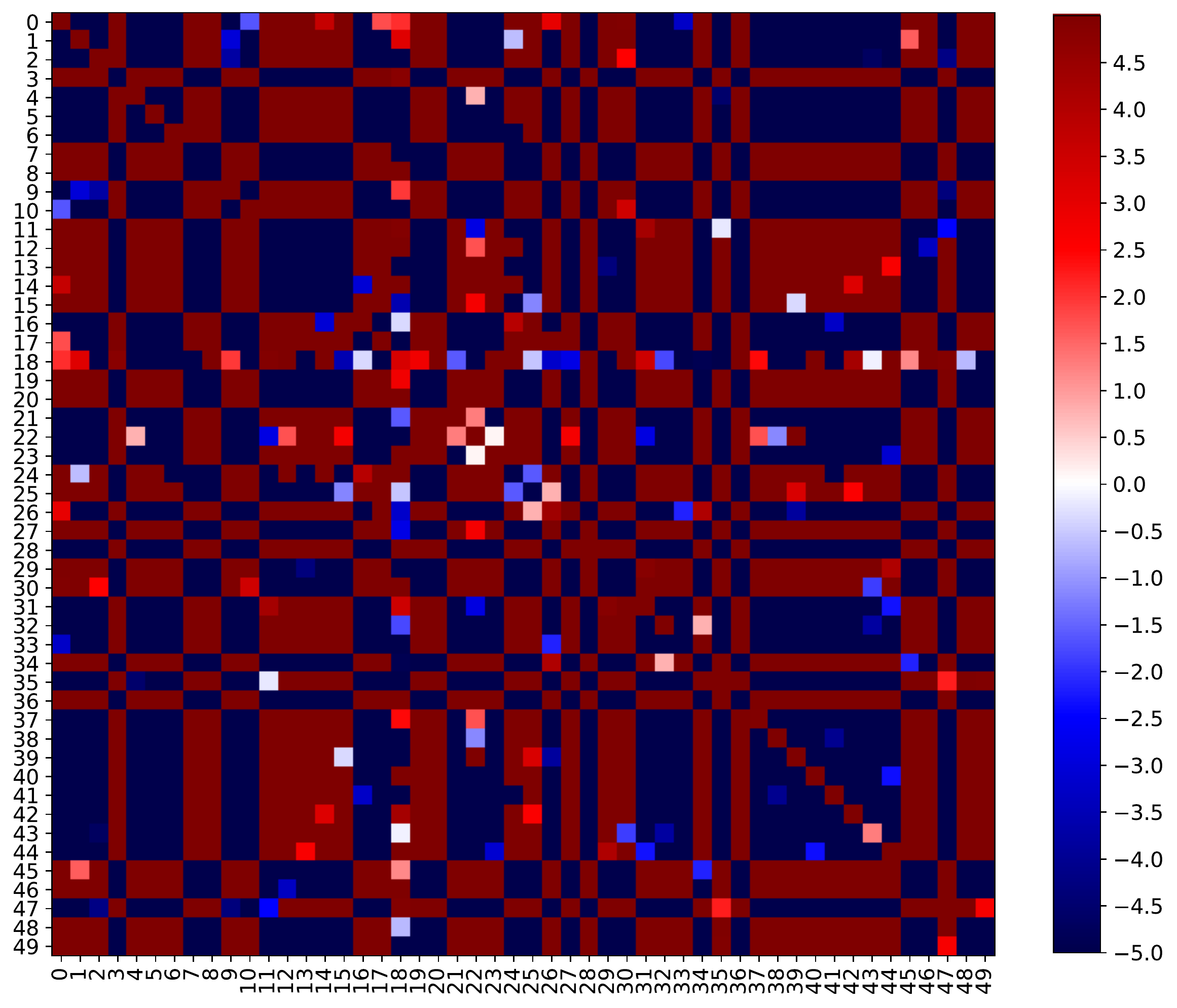}
    \caption{coniii}
      \end{subfigure}
    \caption{Reconstructed Ising models for a system of size 50 with $\beta = 1$ and 75\% sparsity in the matrix elements.} \label{fig:ising_matrix_test} 
\end{figure}
\begin{table}[h!]
  \caption{MSE between the Ising parameters computed using the different methods and ground truth.}
  \label{mse_params}
  \centering
  \begin{tabular}{lll}
    \toprule
    \cmidrule(r){1-2}
    Method     &MSE & $r$    \\
    \midrule
    GNisi     & 14.78  &  0.03 \\
    ACE     & 69.98    & 0.03    \\
    coniii & 408.06 & 0.01\\
    \bottomrule
  \end{tabular}
\end{table}

 In Fig.~\ref{fig:hamiltonian-plot} we plot the scatter of predicted versus ground truth probability distributions, for a sample of possible bit-strings, which are again distinct from the bit-strings used to generate the Ising model for each method. For this model, ACE and GNisi perform very similarly as can also be observed in Table~\ref{log_z} where we show also the Pearson correlation coefficient, $r$.

\begin{table}[h!]
  \caption{Partition functions computed for an Ising model with differing methods}
  \label{log_z}
  \centering
  \begin{tabular}{lll}
    \toprule
    \cmidrule(r){1-3}
    Method     & $\log(Z)$ & $r$     \\
    \midrule
    Ground-truth &  42.86 &$-$    \\
    GNisi     & 46.48  & 0.11   \\
    ACE     & 46.73      &0.10  \\
    coniii & 252.78 & 0.02 \\
    \bottomrule
  \end{tabular}
\end{table}

\begin{figure}[h!]
  \centering
  \begin{subfigure}[h]{0.3\textwidth}
    \centering
    \includegraphics[width=0.9\textwidth]{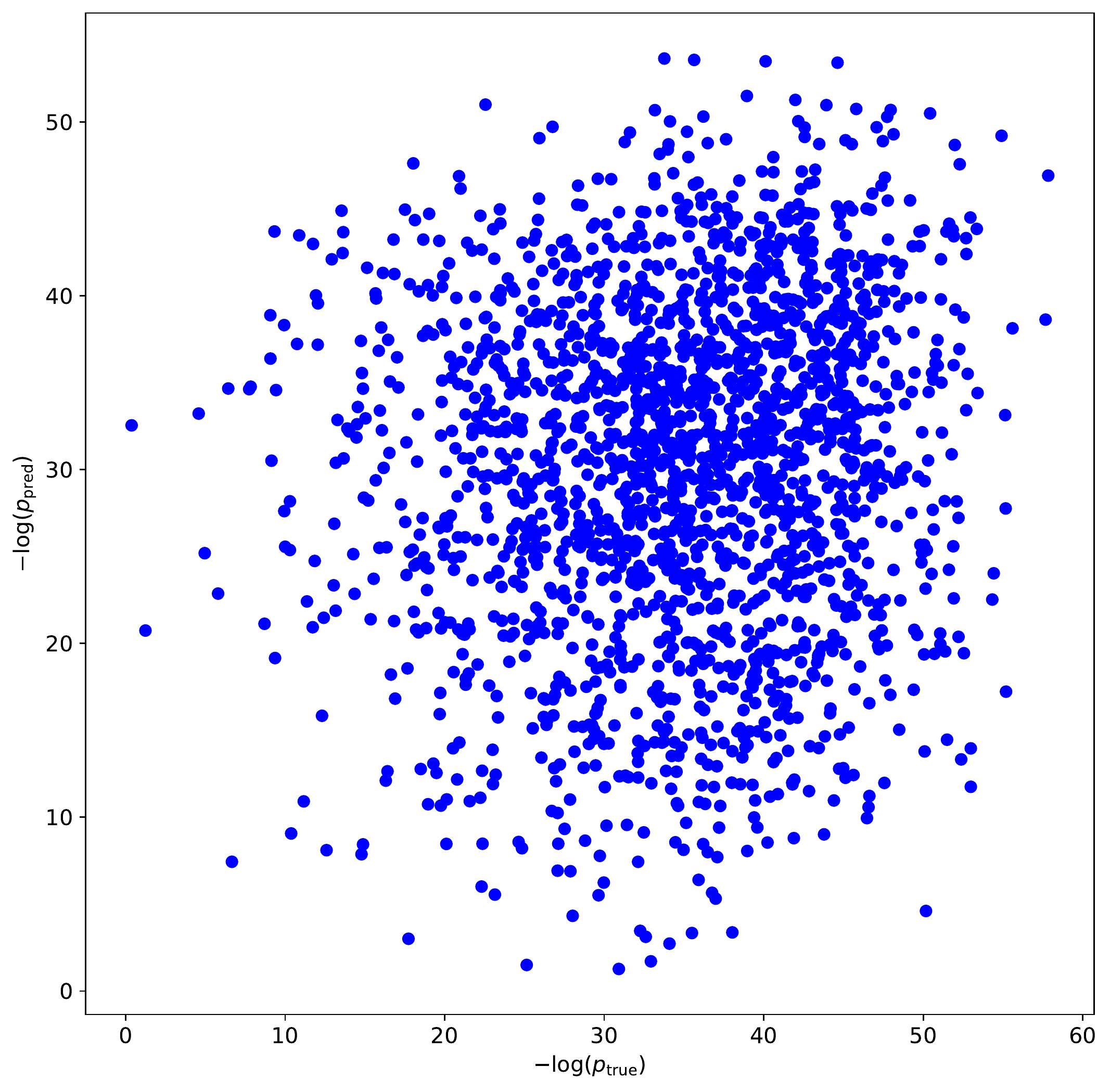}
    \caption{GNisi}
  \end{subfigure}
    \begin{subfigure}[h]{0.3\textwidth}
    \centering
    \includegraphics[width=0.9\textwidth]{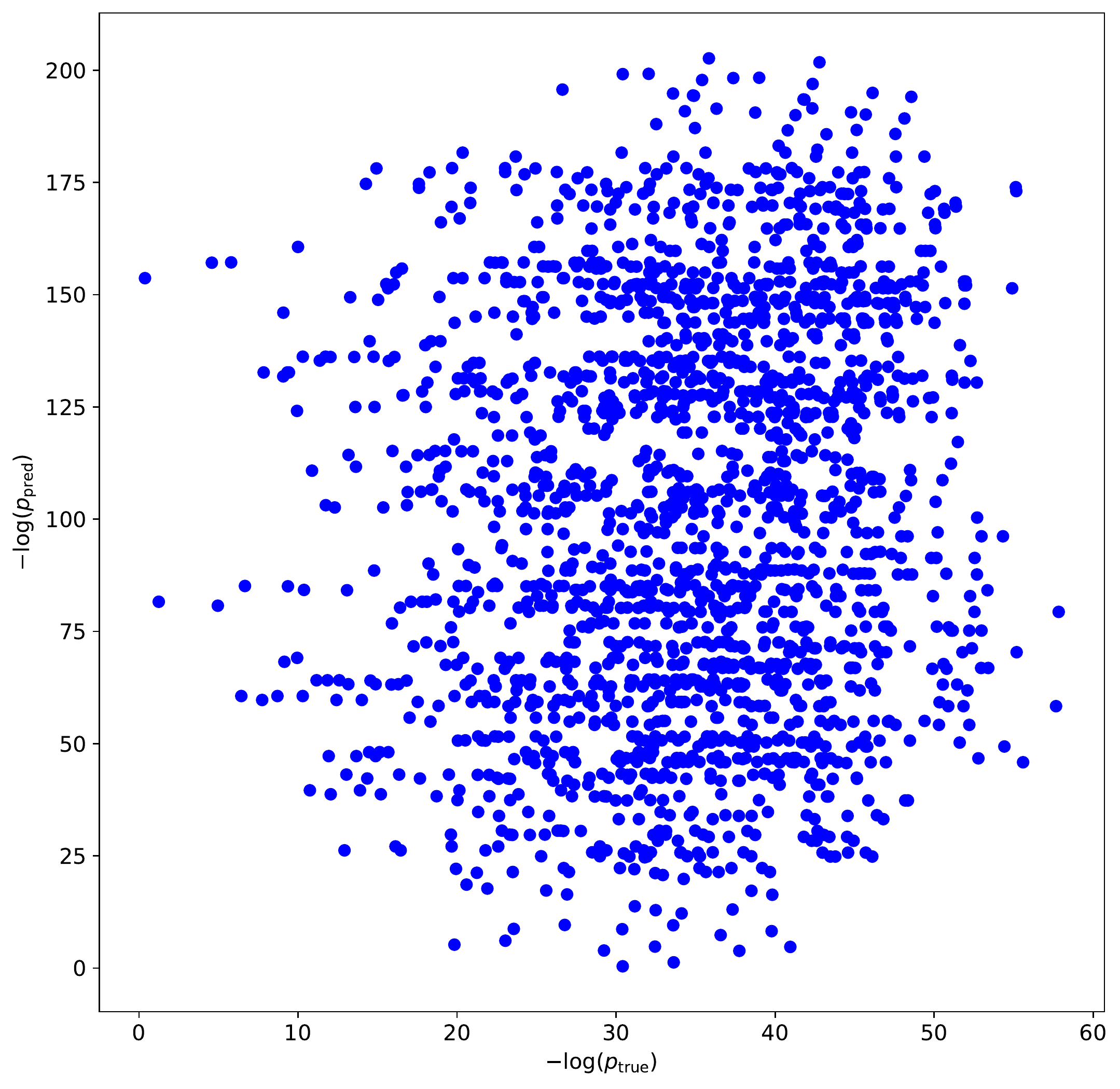}
    \caption{ACE}
  \end{subfigure}
      \begin{subfigure}[h]{0.3\textwidth}
    \centering
    \includegraphics[width=0.9\textwidth]{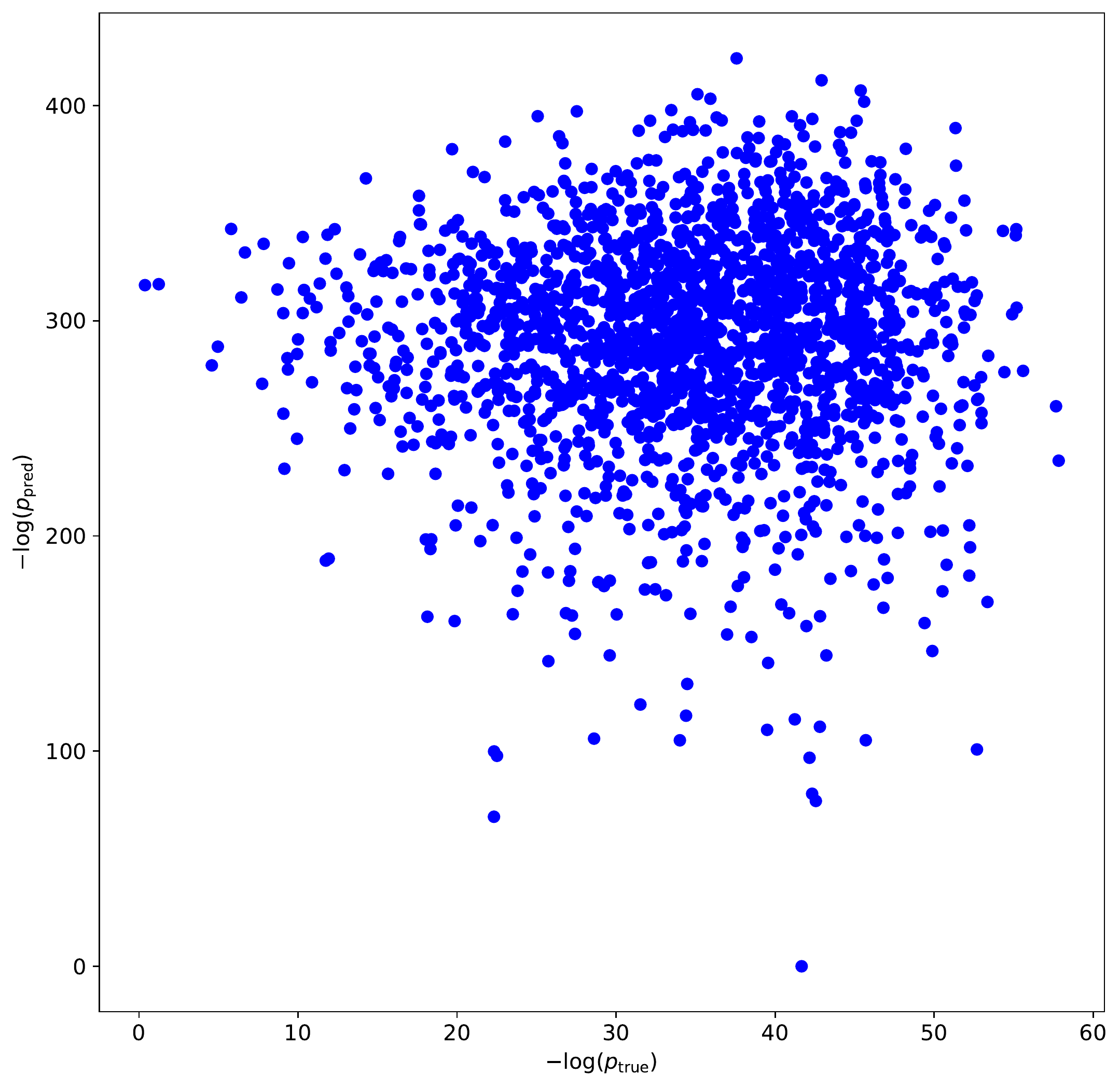}
    \caption{coniii}
  \end{subfigure}
    \caption{Scatter plots of probability distributions computed using the different methods for a sample of possible bit-strings.} \label{fig:hamiltonian-plot}
\end{figure}

Finally we will compare first, second and third moments.
 In Table~\ref{mse-moments-table} we show the mean squared error between the moments computed using the ground truth, and with the various methods. Again, GNisi does very well at matching moments despite it not being a restriction when constructing the Ising model.
\begin{table}
  \caption{Mean squared error between the $n$th moment computed using ground truth and the differing methods. }
  \label{mse-moments-table}
  \centering
  \begin{tabular}{llll}
    \toprule
        \multicolumn{3}{r}{MSE}                   \\
    \cmidrule(r){2-4}
    Moment     & GNisi & ACE  & coniii      \\
    \midrule
    $m_1(1)$     & 0.07   & 0.12 &0.13  \\
     $m_2(1)$   & 1.32$\cdot 10^{-4}$ &    $2.98 \cdot 10^{-7}$&    $ 2.98\cdot 10^{-7}$ \\
          $m_2(0)$   &$1.32 \cdot 10^{-4}$ &    $ 2.98\cdot 10^{-7}$    &$2.98\cdot 10^{-7}$ \\
          $m_3 (1)$   &$2.6 \cdot 10^{-3}$ &   $1.54 \cdot 10^{-6}$ & $1.54\cdot 10^{-6}$ \\
               $m_3 (0)$  & 0.0  &0.0  &0.0\\
    \bottomrule
  \end{tabular}
\end{table}

\section{Accuracy of inverse covariance matrix approximation}
In this section we will compare the accuracy of inverting the covariance matrix over samples compared to ground truth and GNisi. For a larger system of size 50, there are $2^{50} \sim 10^{15}$ equally likely bit-string combinations at high temperature, which is computationally intractable to compute, and one therefore will not expect the inverse covariance matrix computed with 1000 samples to be accurate. We will instead here focus on a much smaller system, of lattice size 5 as the 32 possible combinations of bit strings enable us to perform a fair comparison between GNisi and the inverse covariance matrix. The model shown here is at inverse temperature $\beta=1$ with 55\% sparsity in the matrix elements.

\begin{figure}[h!]
  \centering
  \begin{subfigure}[h]{0.3\textwidth}
    \centering
    \includegraphics[width=0.9\textwidth]{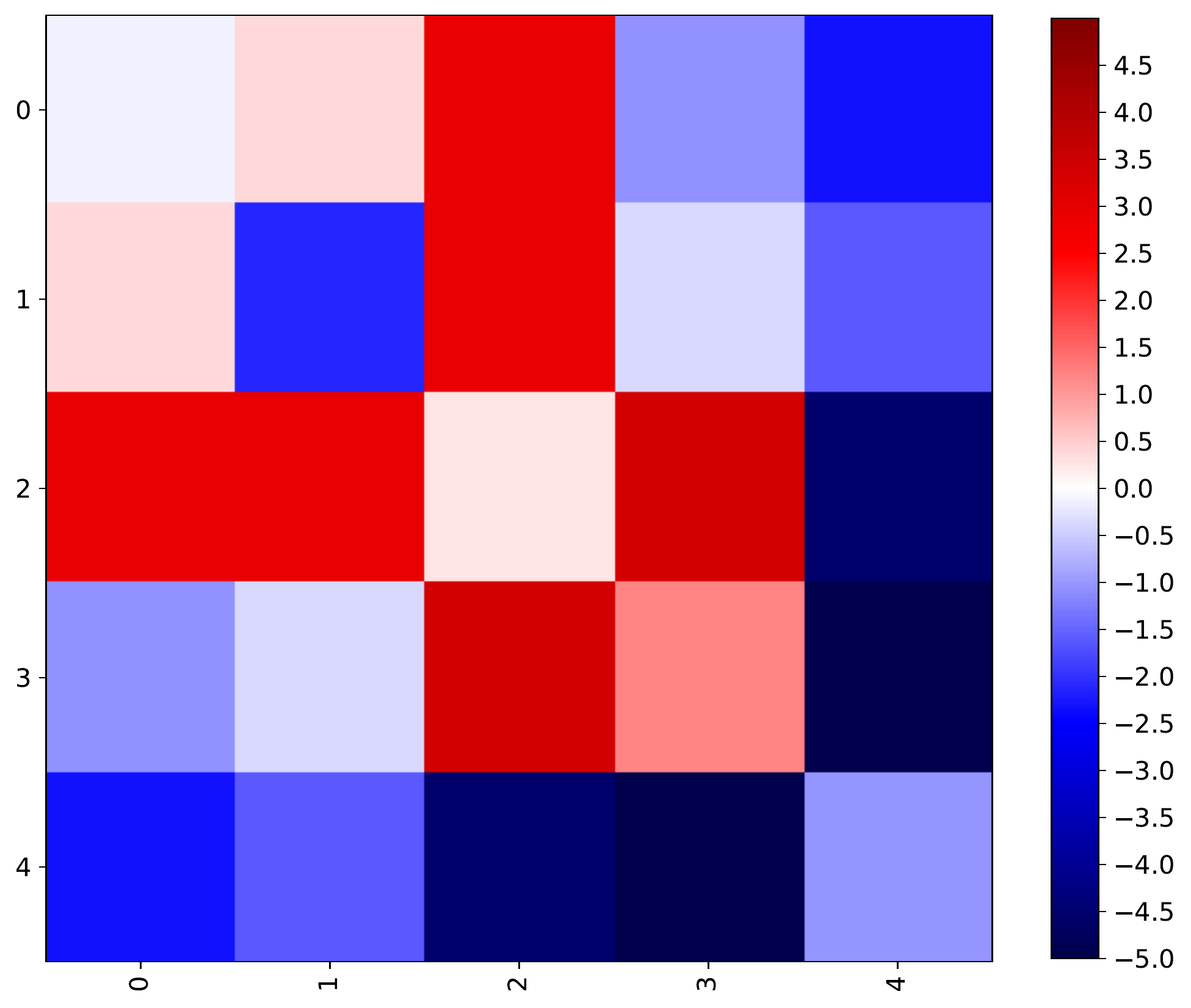}
    \caption{GNisi}
  \end{subfigure}
  \begin{subfigure}[h]{0.3\textwidth}
    \centering
    \includegraphics[width=0.9\textwidth]{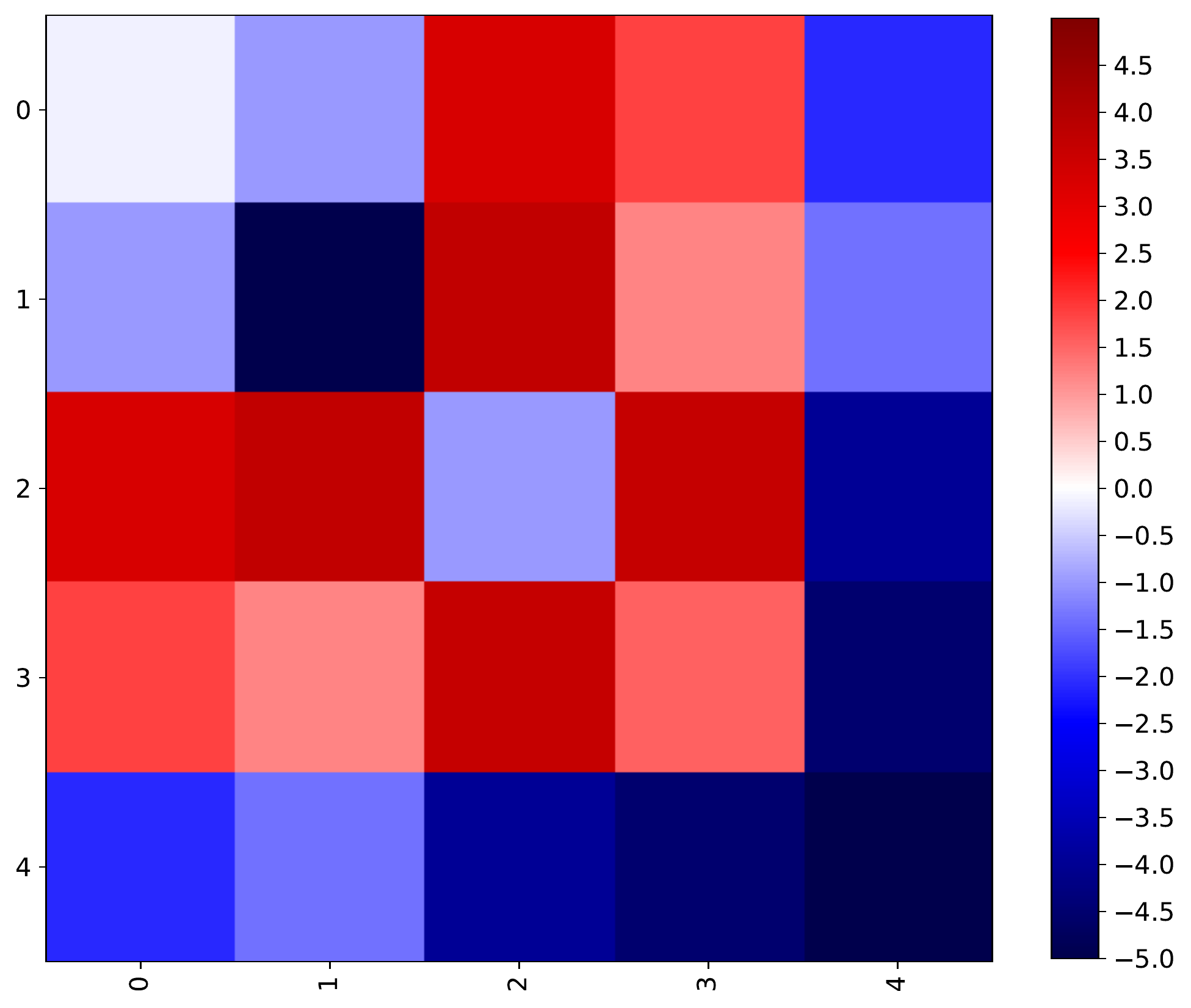}
    \caption{Ground truth}
  \end{subfigure}
   \begin{subfigure}[h]{0.3\textwidth}
    \centering
    \includegraphics[width=0.9\textwidth]{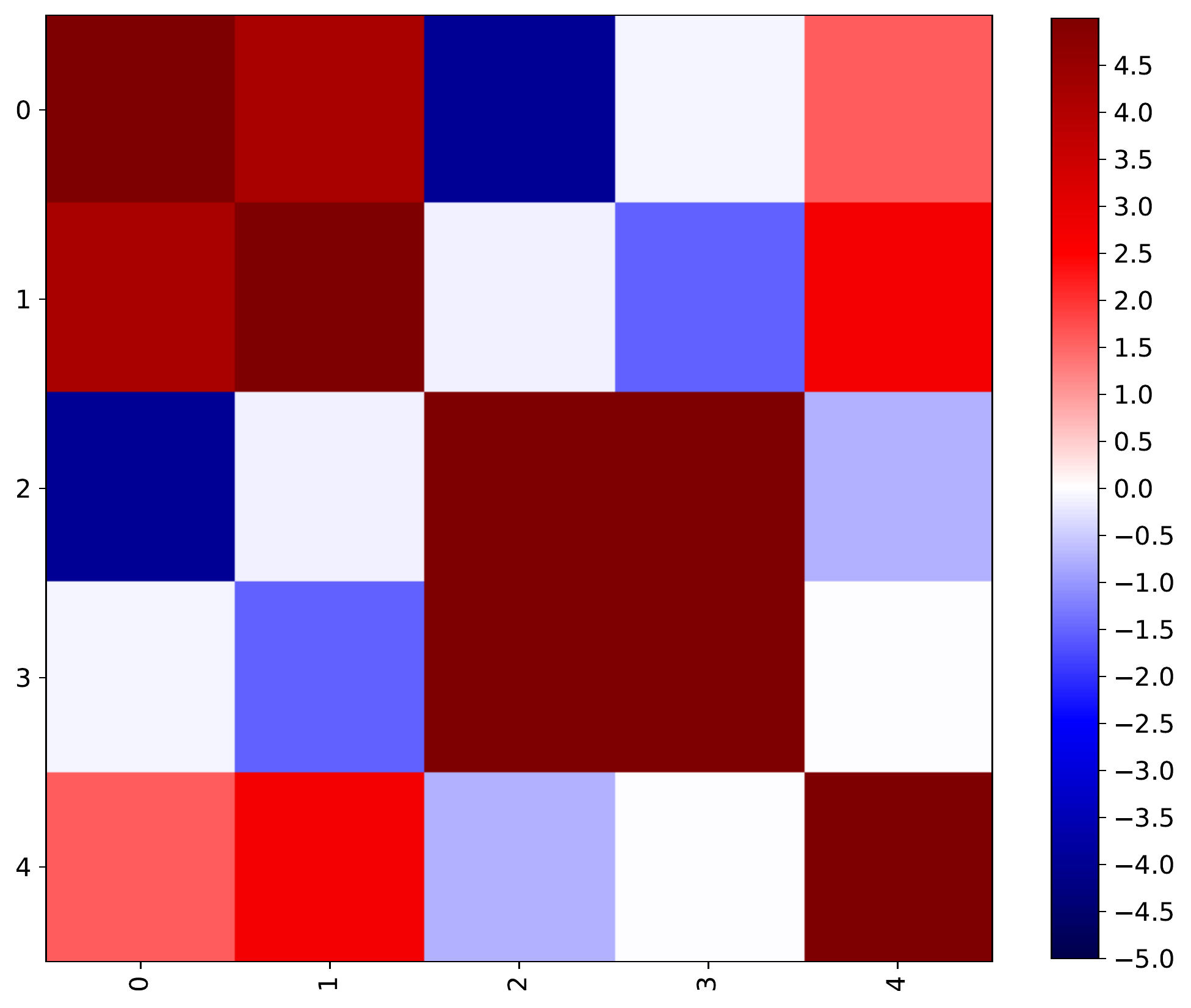}
    \caption{$\text{inv}(\text{cov}_{ij})$}
      \end{subfigure}
    \caption{Reconstructed Ising models for a system of size 5 with $\beta = 1$ and 55\% sparsity in the matrix elements.} \label{fig:ising_matrix_5} 
\end{figure}

We can see in Fig.~\ref{fig:ising_matrix_5} that GNisi matches very well the ground truth, whilst the inverse of the covariance matrix performs very poorly. This is more explicitly shown in  Table~\ref{logz_cov} where we compare the Boltzmann distributions computed using the two methods. We see that GNisi performs well overall, whilst the inverse of the covariance matrix captures nothing about the true distribution.

\begin{table}[h!]
  \caption{Partition functions computed for an Ising model with GNisi and the inverse of the covariance matrix.}
  \label{logz_cov}
  \centering
  \begin{tabular}{ll}
    \toprule
    \cmidrule(r){1-2}
    Method     & $\log(Z)$     \\
    \midrule
    Ground-truth & 19.64    \\
    GNisi     & 12.84     \\
    $\text{inv}(\text{cov}_{ij})$     & 0.00      \\
    \bottomrule
  \end{tabular}
\end{table}

\section{Derivation of maximum entropy approach}
The maximum entropy approach states that, under the condition the first and second moments of the samples match the true connected moments, the Ising model maximises the Shannon entropy of the system. In this section we will derive this result.

Our constraints, in the maximum entropy approximation, are 
\begin{align}
\langle x_i \rangle &= \sum_{\xx} p(\xx) x_i \\
\langle x_i x_j \rangle  &= \sum_{\xx} p(\xx) x_i x_j \\
1 &= \sum_{\xx} p(\xx) \,.
\end{align}
The constrained optimisation problem can be formulated as extremising the Lagrangian of the system, $\mathcal{L}$. We therefore have
\begin{align}
\mathcal{L} &= - \sum_{\xx} p(\xx) \log p(\xx) -  \chi \sum_{\xx} p(\xx) - 1 - \sum_i \left( \alpha_i \left(\sum_{\xx} p(\xx) x_i \right)- \langle x_i \rangle \right) - \sum_{i,j}\left( \gamma_{ij} \sum_{\xx} p(\xx) x_i x_j - \langle x_i x_j \rangle \right) \,. \nonumber 
\end{align}
So
\begin{align}
\frac{\partial \mathcal{L}}{\partial p(\xx)} &= 0 = - \log p(\xx) - 1- \chi  - \sum_i  \alpha_i x_i -  \sum_{i,j} \gamma_{ij}  x_i x_j \,,
\end{align}
and
\begin{align}
 \log p(\xx) = - 1- \chi  -\alpha.\xx- \xx . \gamma . \xx \,,
\end{align}
\begin{align}
 p(\xx) = A e^{-\alpha .\xx- \xx . \gamma . \xx }\,,
\end{align}
which is the Ising model where we identify $\alpha$ as the external field and $\gamma$ as the pairwise terms respectively. 
\section{Derivation of inverse covariance matrix approximation}
In this appendix we derive the approximation where the inverse of the covariance matrix over nodes is approximately equal to the pairwise coupling term. We derive the approximation  using field theoretic methods. 

The partition function for a discrete number of bit-strings is given by the usual
\begin{equation}
Z = \sum_{i=1}^n e^{- \beta H_i(x_i)} \label{eq:partition_function} \,.
\end{equation}
In the infinite temperature limit, all possible $2^n$ combination of spins are equally possible. We can therefore take $i \to \infty$ and then
\begin{equation}
Z = \int \mathcal{D} x e^{- \beta H_i (x_i)}  \label{eq:z_pathint}\,,
\end{equation}
where $\mathcal{D}$ signifies we take path integrals.

The constraints on the first and second moments are now given by
\begin{align}
\langle x_i \rangle &= \int \mathcal{D} x \, p(\xx) x_i \label{eq:moment1} \\
\langle x_i x_j \rangle &= \int \mathcal{D} x \, p(\xx) x _i x_j  \label{eq:moment2}\,,
\end{align}
where 
\begin{equation}
p(\xx) = A e^{-\alpha . \xx - \xx.\gamma . \xx} \,, \label{eq:p_x}
\end{equation}
and $\alpha$ and $\gamma$ are the external field and pairwise terms respectively. Via a change of coordinates $\yy = \xx + \alpha \xi^{-1}$ where $\xi = \frac{1}{2} \gamma$ we can write~\eqref{eq:p_x} as
\begin{equation}
p(\xx) = B e^{ - \frac{1}{2} \yy. \xi \yy} \,, \label{eq:p_y}
\end{equation}
and so~\eqref{eq:moment1} becomes
\begin{align}
\langle x_i \rangle &= \int \mathcal{D} \yy \, B e^{- \frac{1}{2} \yy. \xi . \yy} (y_i - \sum_j \alpha_j \xi^{-1}_{ij}) \\
&= -\int \mathcal{D} \yy \, B e^{ -\frac{1}{2} \yy. \xi . \yy}  \sum_j \alpha_j \xi^{-1}_{ij}\\
& = -B \alpha \xi^{-1} \frac{(2 \pi)^{N/2}}{\sqrt{\text{det} \xi} }\sum_j \alpha_j \xi^{-1}_{ij} \,,
\end{align}
using the fact that
\begin{equation}
\int \mathcal{D} \yy \, B e^{ -\frac{1}{2} \yy. \xi . \yy}   \yy = 0 \,,
\end{equation}
and properties of multi-dimensional Gaussians of size $N$, as we assume in this approximation our data is Gaussian. We therefore have
\begin{equation}
\langle x_i \rangle = -C\sum_j \alpha_j \xi^{-1}_{ij}  \label{eq:moment1_trans} \,. \\
\end{equation}

Eq.~\eqref{eq:moment2} is then
\begin{align}
\langle x_i x_j \rangle &= \int \mathcal{D}  \yy B e^{ - \frac{1}{2} \yy. \xi \yy}  (y_i - \sum_k \alpha_k \xi^{-1}_{ik})(y_i - \sum_l \alpha_l \xi^{-1}_{il}) \\
&= \int \mathcal{D}  \yy B e^{ - \frac{1}{2} \yy. \xi \yy}  (y_i y_j - \sum_l y_i \alpha_l \xi_{jl}^{-1} - \sum_k y_j \alpha_k \xi_{ik}^{-1} +  \sum_{k,l}   \alpha_l \alpha_k \xi_{ik}^{-1} \xi_{jl}^{-1} ) \\
&\sim \int \mathcal{D}  \yy B e^{ - \frac{1}{2} \yy. \xi \yy}  (y_i y_j + y_i \langle x_j \rangle + y_j \langle x_i \rangle +\langle x_i \rangle \langle x_j \rangle) \,,
\end{align}
using Eq.~\eqref{eq:moment1_trans}. Again using the properties of multi-dimensional Gaussians, we find 
\begin{align}
\langle x_i x_j \rangle &= D  \langle x_i \rangle \langle x_j \rangle + \int \mathcal{D}  \yy B e^{ - \frac{1}{2} \yy. \xi \yy} y_i y_j \,. \label{eq:z_functional} \end{align}

The remaining integral is a 2-point correlation function which we can solve analytically. The partition function Eq.~\eqref{eq:z_pathint} is, dropping all constant factors,
\begin{align}
Z & = \int \mathcal{D} \yy e^{-\frac{1}{2} \yy . \xi . \yy + \alpha. \yy}  \\
& = \int \mathcal{D} \xx e^{(\xx + \xi^{-1} \alpha). \frac{1}{2} \xi (\xx + \xi^{-1} \alpha) + \alpha . \xx + \alpha \xi^{-1} \alpha} \\
& \sim   \int \mathcal{D} \xx e^{-\frac{1}{2} \xx \xi \xx + \frac{1}{2} \alpha \xi^{-1} \alpha} \\
& = e^{\frac{1}{2} \alpha \xi^{-1} \alpha} \int \mathcal{D} \xx e^{-\frac{1}{2} \xx \xi \xx } \\
& = e^{\frac{1}{2} \alpha \xi^{-1} \alpha} \frac{(2 \pi)^{N/2}}{\sqrt{\text{det} \xi}} \,.
\end{align}

Now, using the definition of functional derivatives from statistical mechanics
\begin{equation}
\frac{1}{Z[0]} \frac{\delta^2 Z[ \alpha]}{\delta \alpha(x_i)\delta \alpha(x_j)} \bigg|_{ \alpha = 0} := \frac{1}{Z[0]}  \int \mathcal{D} \varphi \varphi(x_i) \varphi(x_j) e^{-S[\varphi]} \,,
\end{equation}
where $S[\varphi]$, the action, can be identified with $H$ and so 
\begin{equation}
\frac{1}{Z[0]} \frac{\delta^2 Z[ \alpha]}{\delta \alpha(x_i)\delta \alpha(x_j)} \bigg|_{ \alpha = 0} = \frac{1}{Z[0]}  \int \mathcal{D} \varphi \varphi(x_i) \varphi(x_j) e^{-H} \,,
\end{equation}
where the integral above is equivalent to the integral in \eqref{eq:z_functional}. Bringing it all together we finally have
\begin{align}
\langle x_i x_j \rangle & \sim \frac{1}{Z[0]} \frac{\delta^2 Z[ \alpha]}{\delta \alpha(x_i)\delta \alpha(x_j)} \bigg|_{ \alpha = 0} + \langle x_i \rangle \langle x_j \rangle \\
& = (2 \pi)^{-N/2} \left[ \text{det}\xi\right]^{\frac{1}{2}}(2 \pi)^{N/2} \left[ \text{det}\xi\right]^{-\frac{1}{2}} \frac{\partial}{\partial \alpha_i}
 \left( \alpha_i \xi_{ij}^{-1} \right) \bigg|_{ \alpha = 0} + \langle x_i \rangle \langle x_j \rangle \\
 & =  \xi_{ij}^{-1} + \langle x_i \rangle \langle x_j \rangle\,. \end{align}

Therefore, the covariance matrix can be identified as being proportional to the inverse pairwise term in the Hamiltonian
\begin{equation}
C_{ij} := \langle x_i x_j \rangle  -  \langle x_i \rangle \langle x_j \rangle \sim \xi_{ij}^{-1} \sim \gamma_{ij}^{-1} \,.
\end{equation}

\section{Definition of the 3rd connected moment}
In this section we state the equation used to compute the 3rd connected moments in the main text. We use the coskewness for $n$-random variables:
\begin{equation}
m_3(i,j,k) = \frac{E[i - E[i]] E[j - E[j]] E[k - E[k]]}{\sigma_i \sigma_j \sigma_k} \,,
\end{equation}
where $E[i]$ is the expected value of $i$ and $\sigma_i$ its standard deviation.

\section{Performance of a random Ising model}
\begin{figure}[!h] 
  \centering
  \begin{subfigure}[h]{0.45\textwidth}
    \centering
    \includegraphics[width=0.9\textwidth]{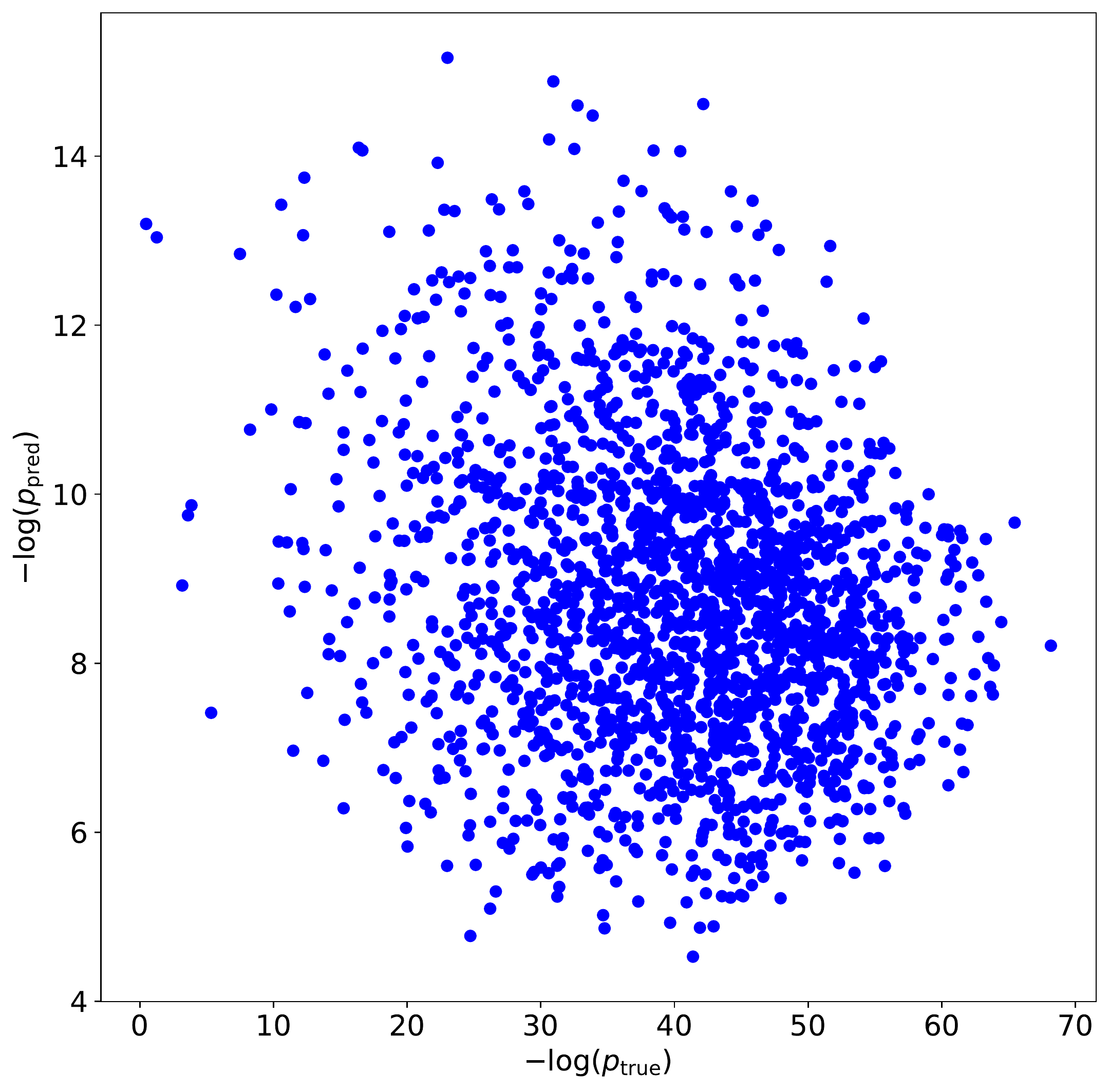}
      \end{subfigure}
        \caption{Scatter plot for the Boltzmann probability distribution for a random matrix against ground truth.} \label{fig:random_matrix}
\end{figure}

In the main text, we state that GNisi, for the Ising model against which we have ground truth, out-performs ACE and coniii. It is reasonable to ask whether a random matrix with parameters of the same size as GNisi would out-perform it. In Fig.~\ref{fig:random_matrix}, we show the scatter plot for the Boltzmann probability distribution for a matrix generated at random with the same size couplings as GNisi. We observe that the probability distribution is captured very poorly; indeed we find $\log(Z) =  7.05$ compared to the ground truth value of $\log(Z) = 55.37$ with the Pearson coefficient on the Boltzmann probability distribution of $r=-0.20$. In Table~\ref{mse-moments-table2} we show the corresponding MSE between the moments computed using ground truth and the random matrix, again observing that GNisi is better. We therefore can conclude that GNisi out-performs a random matrix and is correctly learning the Ising models.

\begin{table}[!h]
  \caption{Mean squared error between the $n$th moment computed using ground truth and the random matrix. }
  \label{mse-moments-table2}
  \centering
  \begin{tabular}{ll}
    \toprule
        \multicolumn{1}{r}{MSE}                   \\
    \cmidrule(r){1-2}
    Moment     & Random matrix       \\
    \midrule
    $m_1(1)$     &   0.10    \\
     $m_2(1)$   & $3.5\cdot 10^{-5}$     \\
          $m_2(0)$   &$3.5 \cdot 10^{-5}$ \\
          $m_3 (0)$   &$1.0 \cdot 10^{-3}$  \\
               $m_3 (1)$  & 0.0\\
    \bottomrule
  \end{tabular}
\end{table}

\end{document}